\documentclass[3p]{elsarticle}

\usepackage{lineno,hyperref}
\usepackage{blindtext, rotating}
\modulolinenumbers[5]
\usepackage{subcaption}

\journal{}

\usepackage{amsmath,amssymb}

\usepackage{changepage}

\usepackage[utf8x]{inputenc}

\usepackage{textcomp,marvosym}
\usepackage{booktabs}
\usepackage{algorithm}
\usepackage{algorithmic}

\usepackage{array}
\usepackage{multirow}

\newcommand\MyBox[2]{
  \fbox{\lower0.75cm
    \vbox to 1.7cm{\vfil
      \hbox to 1.7cm{\hfil\parbox{1.4cm}{#1\\#2}\hfil}
      \vfil}%
  }%
}

\usepackage[table]{xcolor}

\usepackage{array}

\usepackage{multirow}
\usepackage[aboveskip=1pt,labelfont=bf,labelsep=period,justification=raggedright,singlelinecheck=off]{caption}
\renewcommand{\figurename}{Fig}

\usepackage{algorithmic}
\usepackage{rotating}

\biboptions{authoryear}

\begin{document}

\begin{frontmatter}

\title{End-to-End Intelligent Framework for Rockfall Detection}

\author{Thanasis Zoumpekas$^{1*}$}
\ead{thanasis.zoumpekas@ub.edu}

\author{Anna Puig$^1$}
\cortext[mycorrespondingauthor]{Corresponding author}
\ead{annapuig@ub.edu}

\author{Maria Salamó$^1$}
\ead{maria.salamo@ub.edu}

\author{David García-Sellés$^2$}
\ead{dgarcia@ub.edu}
\author{Laura Blanco Nuñez$^{3,4}$}
\ead{lbn@anufra.com}
\author{Marta Guinau$^2$}
\ead{mguinau@ub.edu}

\address{$^1$WAI Research Group, IMUB and UBICS Institutes, Department of Mathematics and Computer Science, University of Barcelona, Barcelona, Spain}
\address{$^2$RISKNAT Research Group, Geomodels Institute, Department of Earth and Ocean Dynamics, University of Barcelona, Barcelona, Spain}
\address{$^3$GGAC Research Group, Geomodels Institute, Department of Earth and Ocean Dynamics, University of Barcelona, Barcelona, Spain}
\address{$^4$Anufra, Soil and Water Consulting, Barcelona, Spain}

\begin{abstract}

Rockfall detection is a crucial procedure in the field of geology, which helps to reduce the associated risks. Currently, geologists identify rockfall events almost manually utilizing point cloud and imagery data obtained from different caption devices such as Terrestrial Laser Scanner or digital cameras. Multi-temporal comparison of the point clouds obtained with these techniques requires a tedious visual inspection to identify rockfall events which implies inaccuracies that depend on several factors such as human expertise and the sensibility of the sensors. This paper addresses this issue and provides an intelligent framework for rockfall event detection for any individual working in the intersection of the geology domain and decision support systems. The development of such an analysis framework poses significant research challenges and justifies intensive experimental analysis. In particular, we propose an intelligent system that utilizes multiple machine learning algorithms to detect rockfall clusters of point cloud data. Due to the extremely imbalanced nature of the problem, a plethora of state-of-the-art resampling techniques accompanied by multiple models and feature selection procedures are being investigated. Various machine learning pipeline combinations have been benchmarked and compared applying well-known metrics to be incorporated into our system. Specifically, we developed statistical and machine learning techniques and applied them to analyze point cloud data extracted from Terrestrial Laser Scanner in two distinct case studies, involving different geological contexts: the basaltic cliff of Castellfollit de la Roca and the conglomerate Montserrat Massif, both located in Spain. Our experimental data suggest that some of the above-mentioned machine learning pipelines can be utilized to detect rockfall incidents on mountain walls, with experimentally proven accuracy.

\end{abstract}


\end{frontmatter}


\section{Introduction}

In the field of geology, one crucial task is rockfall detection, which helps to reduce the risk of future hazards (\cite{Robbins2021GeotechnicalEngineering}). Recently, technologies that are able to characterize the geometrical properties of rock slopes and cliffs have emerged (\cite{Robiati2019ApplicationSlope}). As a result, the possibility of detecting changes in a cliff with high precision has been increased greatly. 

Currently, geoscientists utilize specific methodologies in order to detect changes in rock slopes mainly by comparing measurements at different points in time (\cite{DiFrancesco2020TheClouds}). However, the problem with such techniques is that they rely greatly on the sensitivity of the sensor that captures the data or even the measurement tool. Consequently, due to potential measurement errors, the users need to examine case by case analyzing the nature of the change in the cliff, i.e. vegetation, edge effect, noise or random objects. For instance, it is often difficult to distinguish rock detachments from other changes optically in point cloud or imagery data.

The issues presented above justify the need for a new automated intelligent system designed to detect rockfalls. For this reason, in this paper, we propose an intelligent framework for rockfall detection. At the time of writing, a thorough search for such a system has failed to yield results. In addition, we consider our study to be capable of elucidating the field towards the implementation of a general machine learning classification framework able to handle cases with extremely imbalanced data. The level of data imbalance in this study is significant. This is justified in the later sections of this article since there is a low number of rockfall labelled instances. Depending on the nature of the scenery, the vegetation or edge effects can be numerous compared to rockfall events.

Therefore, our contributions are three-fold: (i) a full intelligent framework for rockfall detection handling imbalanced data; (ii) a study involving real data from two case studies, which shows the efficacy and the efficiency of the developed system; and (iii) a web-based rockfall detection system. Last but not least, our framework could also be incorporated as a component in a geological decision support system. Our study could also be utilized as a baseline methodology and a detection accuracy benchmark for future related experimental analyses. 

In this article, we review several of the intelligent methods dealing with rockfall and landslide detection, we implement the associated machine learning models accompanied by various resampling strategies to handle the imbalanced nature of the study, and apply them on point cloud data to identify rockfall events. Two well differentiated geological environments have been selected as case studies. The first one is located at the Montserrat Massif (Barcelona, Spain), which corresponds to a fractured conglomerate cliff, called Degotalls. The second case study involves data from the basaltic lava flow cliff of Castellfollit de la Roca (Girona, Spain). The acquired data are pre-processed and used as input in the learning phase of our intelligent framework. We systematically compare various models in order to select the most accurate for each case study, to be used in our prototype system for achieving effective rockfall event identifications.

The paper is organized as follows. Section \ref{sec:relat} contains an overview of related efforts and Section~\ref{sec:syste} summarizes the design specifications and the implementation of the developed framework, while also providing the background material required for understanding the whole procedure of detecting rockfall events. In Section \ref{sec:exper}, we present the experimental analysis of the considered intelligent process, based on popular performance metrics for imbalanced classification analysis, and the prototype web-based Rockfall Detection System. Section \ref{sec:final_observations} includes the final observations of our experiments. Finally, Section~\ref{sec:synopsis} contains our concluding remarks and possible issues for further exploration.

\section{Related Work} \label{sec:relat}

Various research studies have been conducted on the intersection of the geology and machine learning domains. Over recent years, a number of machine learning applications have emerged in the geoscience field and there is a growing enthusiasm for intelligent methods~(\cite{Dramsch202070Review}). However, the geoscience domain presents new and special challenges for machine learning algorithms and methodologies, because of the combinations of geoscience properties encountered in each specific case. In addition, there is an open need for novel machine learning research and automatic intelligent analysis, especially in the geosciences field as highlighted by (\cite{Karpatne2019MachineOpportunities}). The automation of intelligent pipelines performing specific, special analysis will bring significant advances in both the geoscience and machine learning domains, especially in the task of detecting objects and events, estimating and long-term forecasting for geoscience variables and in mining relationships in geoscience data.

The detection of deformations and rockfall events have been studied in the literature for a couple of decades (\cite{Crosta2003FailureMeasurements, Carrara1983MultivariateEvaluation}). The number of scientific publications on the use of modern and advanced sensors such as TLS in rockfall and landslide studies has increased considerably in the last years. However, further investigation on the development of new methodologies using such sensors is required in order to address and improve the current time consuming analytical methods due to the high volume of data that these sensors produce (\cite{Abellan2010DetectionMonitoring, Abellan2014TerrestrialInstabilities, Lague2013AccurateN-Z, Oppikofer2009CharacterizationScanning, Jaboyedoff2012UseReview}). Research studies on the \textit{identification} of landslide and rockfall events and on general classification tasks in the geology domain present variations regarding their input data, feature inventories, methodologies and in general their specific objectives.

Regarding the input data, we could categorize the related research studies into three broad categories: (i) the imagery data studies; (ii) point cloud studies; and (iii) other sensory data studies. Considering imagery data studies, remote-sensing techniques are considered to be among the most important for landslide event \textit{detection} and \textit{monitoring}. Multiple efforts by researchers working in the remote sensing field applied in geology appear in the literature, utilizing intelligent methods on high-resolution Earth Observation imagery data~(\cite{Prakash2020MappingModels}). Imagery data seem to be a popular data source utilized in numerous studies. Recently published articles, such as (\cite{Xiao2018LandslideHighway, Fanos2018AData, Pham2018AIndia, Bandura2019MachineApplications, Ghorbanzadeh2019EvaluationDetection, Wang2020LandslideLearning, Bui2020ComparingAssessment}) use images as their main data source. In addition, researchers involved in (\cite{James2012StraightforwardApplication}) constructed 3D images using a multi-view stereo algorithm and consumer-grade cameras. 

On the other hand, the use of TLS devices to extract point cloud data for the categorization of landslide kinematics is significant. Also, research studies show that it seems to be necessary to integrate point cloud data from various sources, namely TLS combined with Airborne Laser Scanner (ALS) or Structure from Motion (SfM) methods combined with TLS, in order to overcome the limitations of each individual technique~(\cite{Lissak2020RemoteHazards}). Point clouds are heavily utilized by various researchers, mainly because they carry detailed, high quality information. Specifically, recent studies, including (\cite{Mayr2018MultitemporalMonitoring, Weidner2019ClassificationAnalysis, Weidner2020GeneralizationClassifiers, Kong2020AutomaticClouds, HuuPhuong2020LidarAlgorithm, Fanos2020MachineGIS, Weidner2020AutomatedLearning, Wang2020AutomaticSlopes}) use point clouds as their main data source for their analysis. Moreover, advanced research studies use both images and point clouds portraying a speciality and differentiate themselves from the majority of the related studies (\cite{Fanos2019AGIS, Bernsteiner2020MachineImagery, Loghin2020SupervisedLearning}). Furthermore, \cite{Salvini2013PhotogrammetryAlps} use images and TLS point cloud data to model slopes. Last but not least, \cite{Hemalatha2019EffectiveLearning} utilize data from wireless sensor networks and machine learning algorithms in order to monitor and forecast landslides in real time.

Several researchers have compiled a brief selection of the features used for landslide and rockfall identification. These features can be categorized in five groups, namely morphological, hydrological, geological, land cover features and features obtained from other sources, like rainfall intensity, according to \cite{Prakash2020MappingModels}. Pure geological features like the lithology of a studied field or geo-structural information appear in older studies such as (\cite{Agliardi2001StructuralKinematics}). Recently, various investigators have used explanatory variables obtained from simple cartographic operations on geological data. Commonly, these variables form land cover features determined from distances to faults, rivers, and roads, which are calculated using basic spatial operations in the GIS software (\cite{Reichenbach2018AModels}). In addition, \cite{Wang2020LandslideLearning} use combined morphological predictors, namely elevation, aspect, curvature, slope, with hydrological ones like wetness index and rainfall intensity data. Similarly, \cite{Micheletti2014MachineMapping} utilize morphological and hydrological features adding multiple curvature measurements, such as plan and surface curvature in order to explain the mechanisms of landslides and to confirm the state of vegetation, roads and, in general, geometric deformations.

Furthermore, there are variations regarding the methodologies utilized in rockfall and landslide \textit{identification} and \textit{forecasting}, ranging from methods using traditional geology and statistics to more intelligent ones using machine learning approaches such as deep learning and neural networks. Multiple efforts have been presented introducing analysis frameworks able to deal with landslide and rockfall events.

Traditional geology is mainly utilized in research articles published more than 10 years ago, such as (\cite{Abellan2009DetectionEvent}) although recent papers still investigate the application of such methods (\cite{Robiati2019ApplicationSlope}). \cite{Lague2013AccurateN-Z} proposed a method, which compares and combines two different sets of point clouds obtained from a TLS device. According to their methodology, two different time frames of point cloud data are combined and later compared in order to identify whether the input data form deformation or rockfall event. The aforementioned method uses traditional statistics and lacks of generalization performance.

Machine learning based methodological workflows are the most used techniques, as there are various research articles utilizing them. Recent research studies such as (\cite{Weidner2019ClassificationAnalysis, Hemalatha2019EffectiveLearning, Weidner2020GeneralizationClassifiers, Kong2020AutomaticClouds, HuuPhuong2020LidarAlgorithm, Bernsteiner2020MachineImagery, Loghin2020SupervisedLearning, Weidner2020AutomatedLearning, Wang2020AutomaticSlopes}) utilize mainly machine learning algorithms for rock-slope and landslide \textit{monitoring} and \textit{analysis}. More elaborate machine learning methods outperform the majority, examples include \cite{Mayr2017Object-basedMonitoring}, who developed a landslide monitoring approach for TLS point cloud data, which integrates a specialized machine learning classification with topological rules in an object-based analysis framework. 

Deep learning and neural networks are currently utilized in fewer studies, such as (\cite{Shi2019GeologyMethods, Chen2015DeformationNetwork}). However, advanced neural network architectures using specialized loss functions are being employed to tackle specific issues in the field such as data imbalance in landslide analysis, showing important advances in terms of performance~(\cite{Prakash2020MappingModels}). Also, \cite{Xiao2018LandslideHighway} present a landslide susceptibility assessment framework based on deep learning algorithms using multi-source imagery data. A wide assortment of recently published research studies combines and compares machine learning, neural networks and ensemble based techniques in order to achieve higher performance in terms of accuracy and generalisability (\cite{Bui2020ComparingAssessment, Xiao2018LandslideHighway, Fanos2018AData, Fanos2019AGIS, Fanos2020MachineGIS,Pham2018AIndia,Ghorbanzadeh2019EvaluationDetection, Wang2019FrustumAmodal, Bandura2019MachineApplications, Wang2020LandslideLearning, Fanos2020MachineGIS}).

Landslide and rockfall \textit{identification} are two tasks that present significant imbalanced data issues. The data imbalanced issue has been addressed in multiple studies. \cite{Prakash2020MappingModels} approach the imbalanced learning task by utilizing sophisticated loss functions in the training phase and data augmentation techniques. \cite{Stumpf2011Object-orientedForests} proposed a machine learning based analysis framework with an iterative scheme to handle class imbalance. \cite{Zhao2020UsingSusceptibility} tackled minor data imbalance issues in predicting landslide susceptibility by utilizing a voting system and the random processing of samples with a random forest algorithm.

In a more recent framework, \cite{Fanos2020MachineGIS} proposed combining ALS and TLS data with GIS and introducing a hybrid ensemble model and a 3D kinematic rockfall forecasting model able to deal with rockfall hazard assessment, achieving promising accuracy. However, they did not address the imbalanced nature of the study or deal with imbalanced data. Our proposal considers a different machine learning approach that also deals with the problem of imbalanced data using different types of data sources. Specifically, we propose an intelligent analysis framework and rockfall detection decision support software, that incorporate various processing stages considered to be essential for such problems, namely clustering, resampling, model parameterization and feature selection. Regarding the input data sources, we consider only point cloud data acquired by a TLS device from two distinct geological contexts, with features derived mostly from the morphology of the terrain.

\section{End-to-End Machine Learning Framework: Design and Implementation} \label{sec:syste}

This section introduces our developed end-to-end machine learning framework for detecting rockfall events. First, we explain the basic procedure of detecting rockfalls. Then, we analytically present our proposed framework.

\subsection{Background on Detecting Rockfalls}
\label{sec:basic_proc}
\begin{figure}
        \centering
        \includegraphics[width = \textwidth]{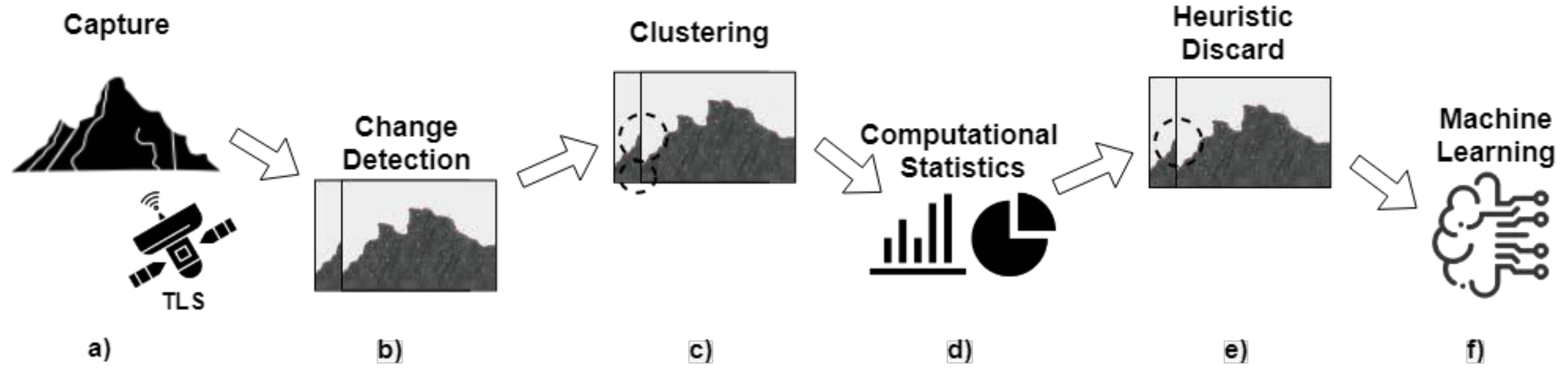}
        \caption{Procedure for detecting rockfalls.}
        \label{fig:process_overview}
\end{figure}

One of the most common processes in the detection of rockfall events on mountain cliffs and slopes is explained in (\cite{DiFrancesco2020TheClouds}) and depicted in Fig~\ref{fig:process_overview}, specifically in steps (a) - (e). The methodology implies capturing periodical measurements of the same cliff face at different time-frames with a Terrestrial Laser Scanner (TLS), denoted as step (a) in Fig~\ref{fig:process_overview}. TLS is a measuring device which offers the ability to collect dense point-clouds of objects. It also provides high-precision and high-accuracy data and is widely used in the geology domain~(\cite{Williams2011MonitoringMapping}). After the capture, the procedure continues with the detection of changes in the surface of the cliff from point cloud data comparison, which is mostly performed with a technique called \textit{m3c2}~(\cite{Lague2013AccurateN-Z}), denoted as step (b). As a result, a new point cloud is obtained containing the metric distances between the compared dense clouds. A clustering algorithm is then applied in step (c) following the change detection step outputting a set of clusters, which are aggregations of points with a significant distance value. These clusters allow for the management of subsets of point clouds with specific topological properties. A statistical analysis then generates a set of features (step (d)) to be evaluated manually, through a visual inspection with expert criteria, to determine whether they are rockfalls or random noise (step (e)). In most cases, samples, i.e. the clusters, happen to be random noise or even measurement errors by the scanner itself.

The current semi-manual classification task of the TLS point cloud data for the detection of rockfall events presents two main challenges. The first concerns the sensibility of the sensor because in some cases the detection of movements is smaller than the margin of error of the device. The second concerns the process of distinguishing rock movement events from other kinds of events, such as the movement of the sensor between measurements, the appearance of vegetation or even random noise. The use of clustering techniques helps to mitigate the aforementioned issues.

Machine learning algorithms present an interesting and currently widely accepted solution for the automatic classification of TLS point cloud data for the detection of rockfall events. However, rockfall detection is considered to be a highly imbalanced classification task, due to the rarity of a rockfall event in a relative dataset. Moreover, clustered point cloud data present patterns that are not easily distinguishable, while having high dimensionality, due to the considerable number of features. In addition, automating the above-mentioned process seems to be an interesting solution for an individual working in the geology domain. There is an open need for a consistent and concise intelligent rockfall identification framework, as detailed by (\cite{Karpatne2019MachineOpportunities}). In Section \ref{sec:system_overview}, we propose a developed machine learning framework, which represents step (f) in Fig~\ref{fig:process_overview}.

\subsection{Framework Overview} 
\label{sec:system_overview}

\begin{figure}
    \centering
    \includegraphics[width = \textwidth]{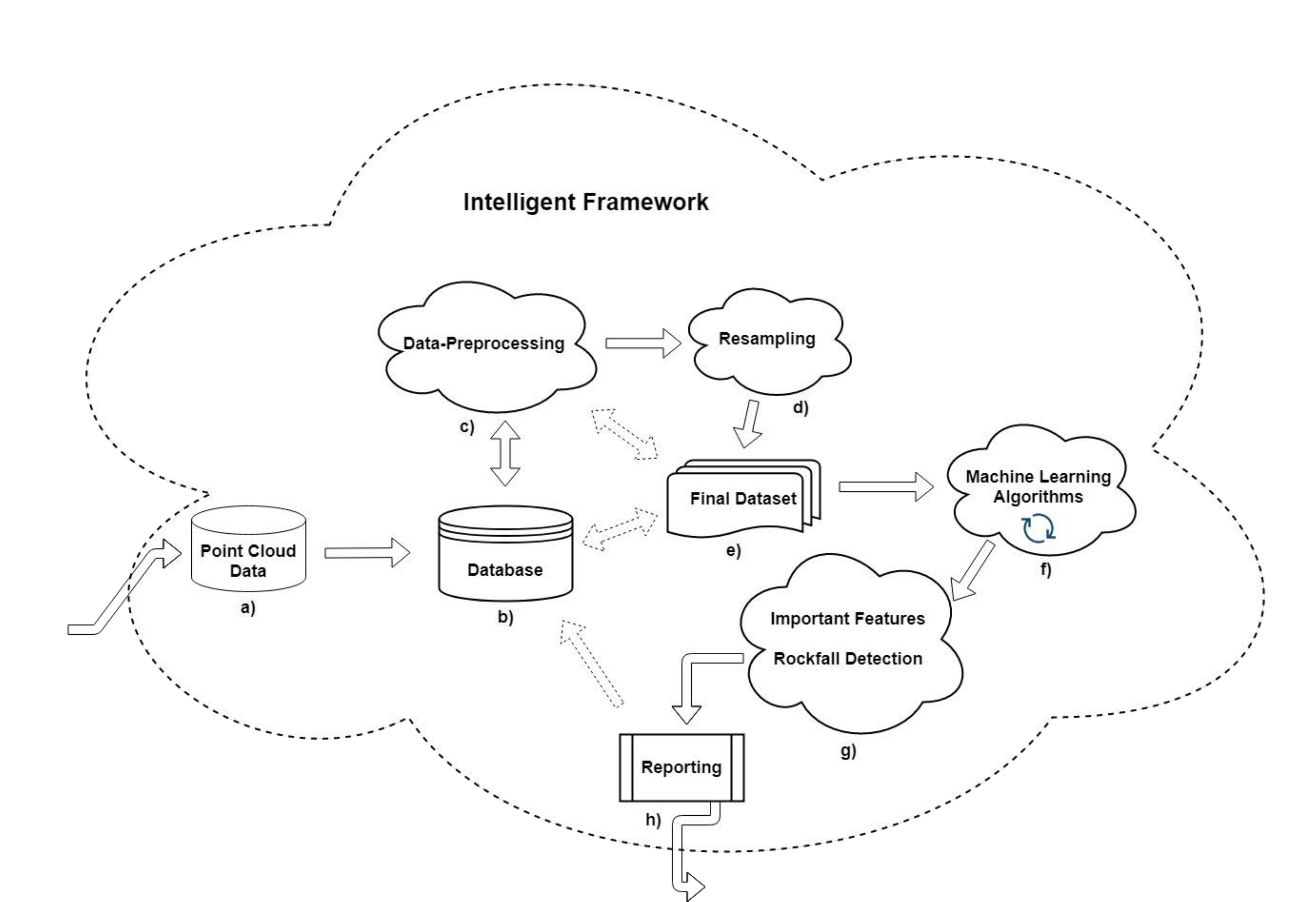}
    \caption{Component-based system architecture.}
    \label{fig:sys_arch}
\end{figure}

\begin{figure}
    \centering
    \includegraphics[scale = 0.27]{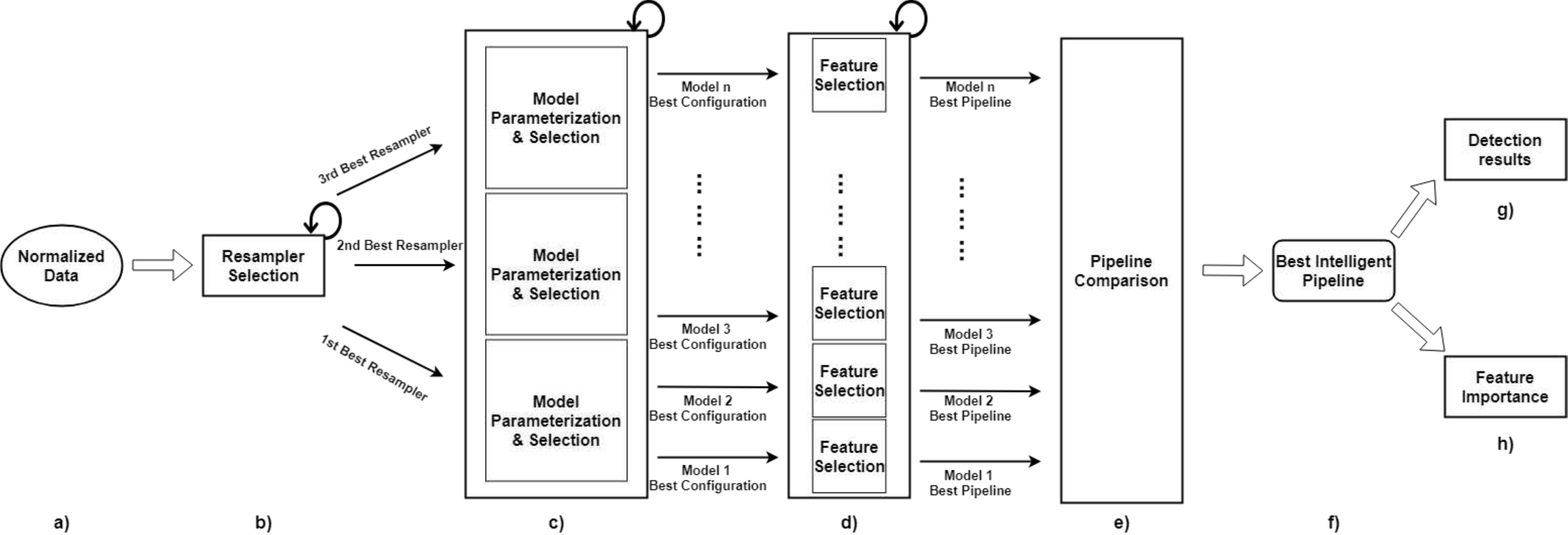}
    \caption{Intelligent framework - methodology workflow.}
    \label{fig:methodology}
\end{figure}

An outline of our proposal can be seen in Fig~\ref{fig:methodology}, where we synthesize our framework by providing its methodology workflow, and in Fig~\ref{fig:sys_arch}, where we present its component-based system architecture.

First, we collect point cloud data in different time frames (Fig \ref{fig:sys_arch} (a)) using the TLS device and store it in a database (Fig \ref{fig:sys_arch} (b)). Next, we pre-process it (Fig \ref{fig:sys_arch} (c)) in order to be fed in our analysis framework. Then, we statistically analyse the input data, obtaining descriptive statistics to elucidate the field regarding the data. The obtained dataset is then stored and normalized. Subsequently, to prevent issues of imbalance, we define our resampling module, portrayed in Fig \ref{fig:sys_arch} (d), which balances the normalized data (Fig \ref{fig:sys_arch} (e)) to be used as input for machine learning algorithms (Fig \ref{fig:sys_arch} (f)). Several intelligent machine learning pipelines are utilized to detect rockfall events and to identify the most significant features (Fig \ref{fig:sys_arch} (g)). Finally, a reporting component is used in order to display the results in a convenient and illustrative manner, as depicted in Fig \ref{fig:sys_arch} (h).

Regarding our methodology, we initially feed normalized point cloud data in our intelligent framework as depicted in Fig \ref{fig:methodology} (a). Then, in the resampling process (Fig \ref{fig:methodology} (b)), an assortment of several resampling strategies accompanied by multiple machine learning models are investigated in order to handle this imbalanced classification task. The three best resamplers are then selected and paired with a variety of models in the model selection and parametrization stage (Fig \ref{fig:methodology} (c)). The completion of the hyper-parameterization stage follows the feature selection phase (Fig \ref{fig:methodology} (d)), in which the best parameterized models of each model variant accompanied by each one of the three best resamplers are experimentally evaluated using different numbers of features. The output of the aforementioned stage consists of multiple properly parameterized machine learning pipelines (Fig \ref{fig:methodology} (e)). These pipelines are then statistically compared in order to decide on the best intelligent pipeline to be used in our prototype system (Fig \ref{fig:methodology} (f)), which is able to produce detection results (Fig \ref{fig:methodology} (g)) and rank feature importance (Fig \ref{fig:methodology} (h)) with promising accuracy. In the following sub-sections, we analytically present our framework and methodological approach.

\subsection{Data Collection}
\label{subsec:data_collection}

The data used in this study has been freely provided by the RISKNAT\footnote{\href{http://www.ub.edu/risknat/}{http://www.ub.edu/risknat/}} research group and the GEOMODELS\footnote{\href{http://www.ub.edu/geomodels/}{http://www.ub.edu/geomodels/}} Research Institute, which belong to the University of Barcelona. 

In this article, we utilize two distinct case studies. Regarding the first case study, we use point cloud data measured from the Degotalls cliff in Montserrat Massif, located in Barcelona, Spain. The dataset consists of clustered TLS point cloud data aggregated from temporal point cloud measurements, from 2007 to 2020 in eight nearly regular time steps. The second case study includes clustered TLS point cloud data measured from the cliff of Castellfollit de la Roca, located in Girona, Spain, from 2008 and 2012. We would like to highlight that the two landscapes are very different in terms of their geological and imagery properties.

Actually, in each case study, the data is composed of a set of statistics computed from the point cloud data, using a technique based on the \textit{m3c2} methodology~(\cite{Lague2013AccurateN-Z}), which creates clusters of points, as described in Section \ref{sec:basic_proc}, including several numerical features that are used in the proposed workflow to identify rockfalls. Then, the aforementioned statistical set is fed in our analysis framework. The Degotalls data consist of 6004 instances, i.e. clusters, and has 37 distinct numerical features, while the Castellfollit data consist of 10371 instances and has 31 numerical features. The aforementioned features include the coordinates of the point clouds and other statistically computed values. A list of these variables with a brief explanation can be found in \nameref{S1_Appendix}.

\subsection{Data Analysis}
\label{sec:data_analysis}

This section provides an exploratory data analysis of the point cloud data. This analysis is fundamental in predictive analytics because it reveals not only the existence of missing data but also elucidates the landscape of the whole data collection phase. Specifically, in tables~\ref{tbl:summary_degotalls} and \ref{tbl:summary_castell} in \nameref{S2_Appendix}, we present the summary statistics for all the data features used in the Degotalls and Castellfollit case studies respectively, providing basic information for our subsequent analysis. For clarification purposes, we chose to use an abbreviation for each variable of the data. The complete variable names followed by a brief explanation can be found in \nameref{S1_Appendix}.

The classification labels of the clusters are determined by the event that causes the change in the surface of the cliff. Rockfall events are denoted as "Candidate". Also, we call "Precursor" the events in which the rock presents a small movement prior to rockfall. In addition, the vegetation of the cliffs is denoted as "Vegetation" and the unknown or human-based events as "Unknow". The artifacts due to the edge effects or the TLS noise are denoted as "Limit\_effect".

Observing the graphs in Fig \ref{fig:cluster_labels}, it is clear that the data in each case study are considered highly imbalanced. In the Degotalls case study (Fig \ref{fig:cluster_labels} (a)), we have only 65 rockfall candidate samples compared to the total of 6004 samples. On the other hand, the Castellfollit case study (Fig \ref{fig:cluster_labels} (c)) includes only 38 compared to the total of 10371 samples. Please note, that the labelling process is done by expert geoscientists with a visual inspection of the 6004 and the 10371 clusters.

\subsection{Data Preprocessing} 
\label{sec:data_prep}

The data analysis in Section \ref{sec:data_analysis} shows that the input data contain values of a varying range of significant size, so normalization should be conducted in order to assist the optimization algorithms to converge faster and the machine learning models to achieve the best performance~(\cite{Singh2019InvestigatingPerformance}). We first normalize the data using z-score normalization technique as proposed in~(\cite{Singh2019InvestigatingPerformance}), which is defined in equation \ref{eq:norm}.
\begin{equation}
\label{eq:norm}
    X_{norm} = \frac{X - mean(X)}{std(X)},
\end{equation}

\noindent where $X$ denotes the feature array and $X_{norm}$ is the normalized $X$ as resulting from the subtraction of its mean ($mean(X)$) and division by its standard deviation ($std(X)$). 

\begin{figure}
    \centering
    \begin{subfigure}[t]{0.45\textwidth}
        \centering
        \includegraphics[width = \textwidth]{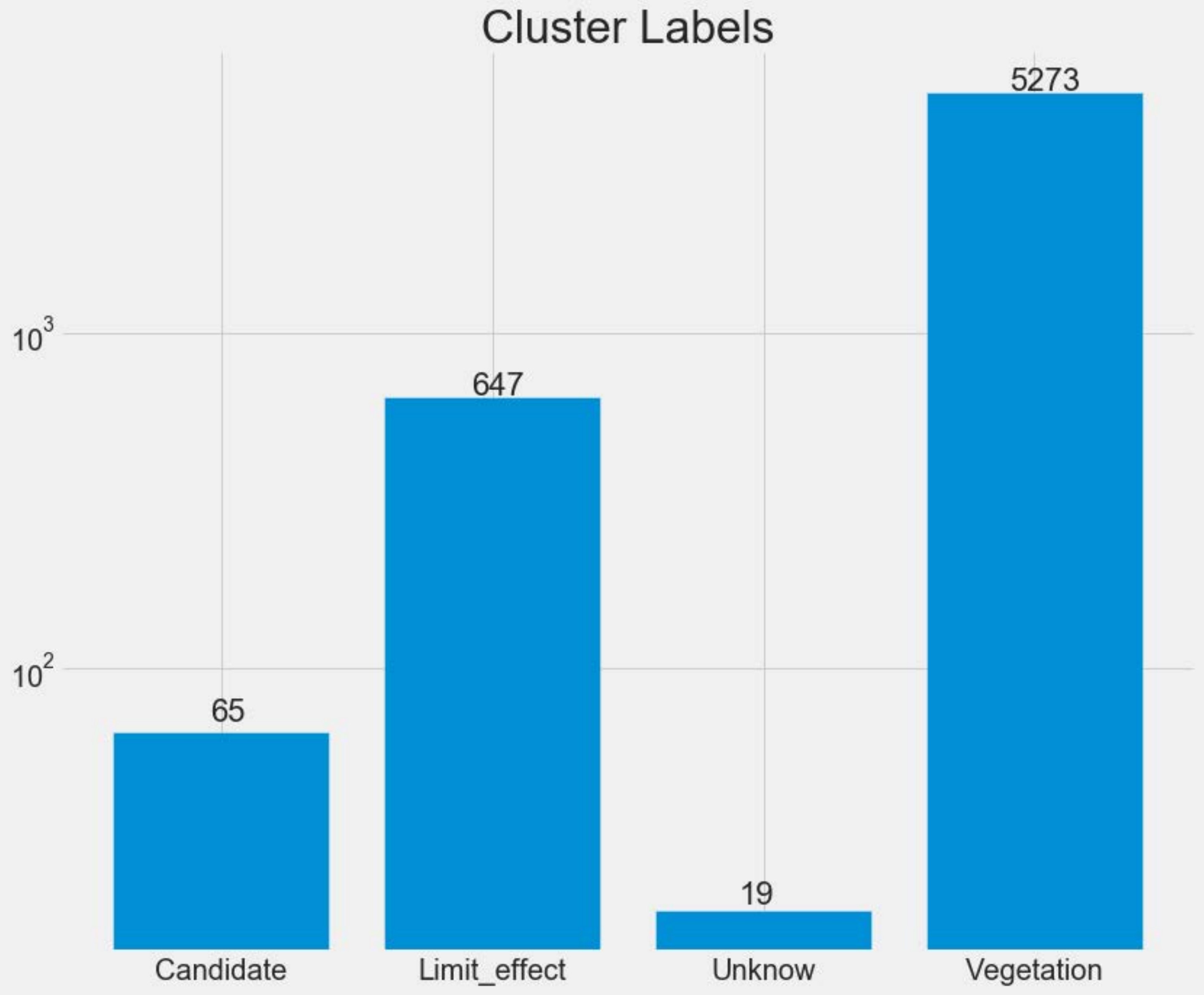}
        \caption{Cluster labels - Degotalls case study.}
    \end{subfigure}%
    ~ 
    \begin{subfigure}[t]{0.45\textwidth}
        \centering
        \includegraphics[width = \textwidth]{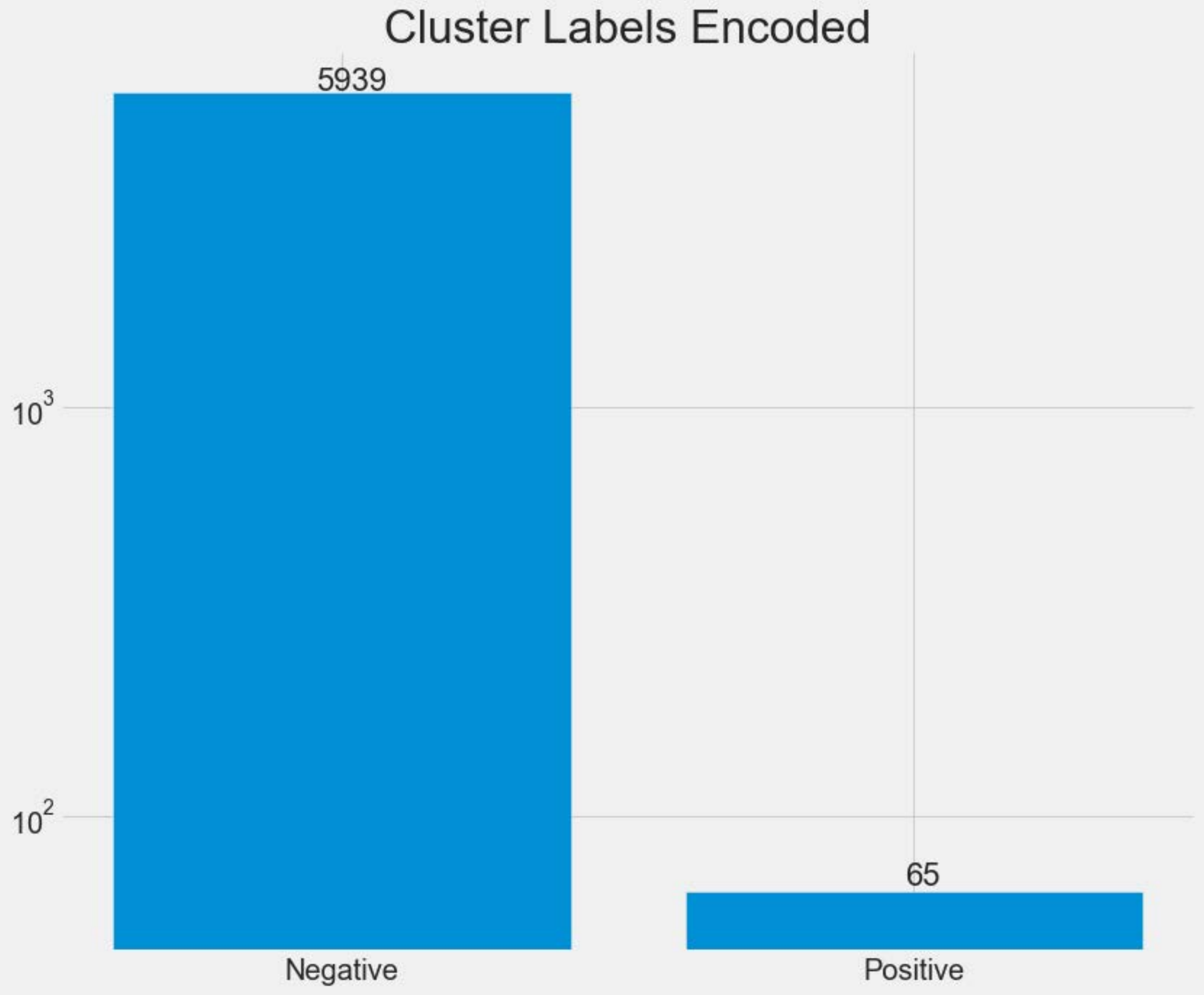}
        \caption{Encoded cluster labels for classification task - Degotalls case study.}
    \end{subfigure}
    ~ 
    \begin{subfigure}[t]{0.45\textwidth}
        \centering
        \includegraphics[width = \textwidth]{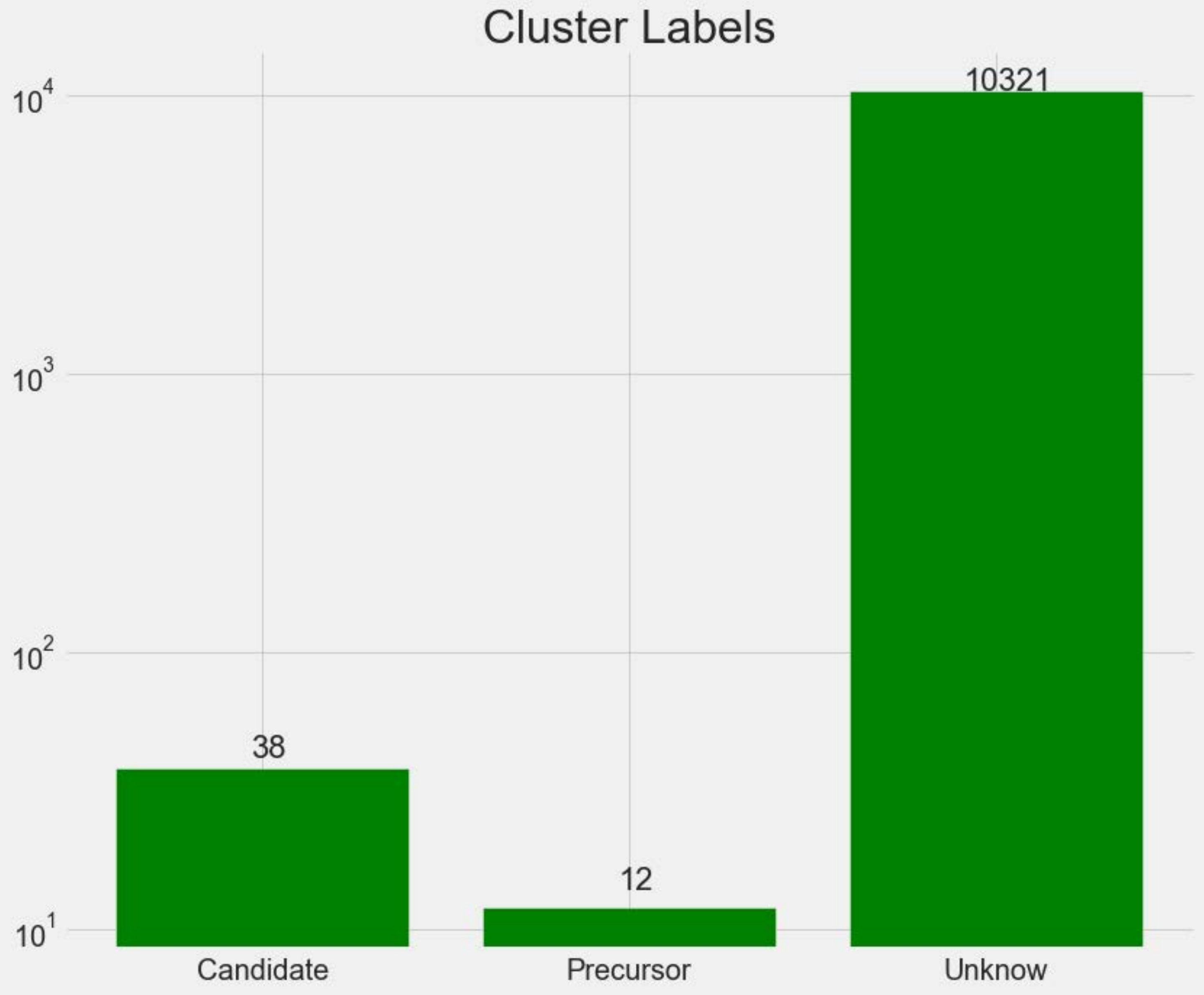}
        \caption{Cluster labels - Castellfollit case study.}
    \end{subfigure}
    ~ 
    \begin{subfigure}[t]{0.45\textwidth}
        \centering
        \includegraphics[width = \textwidth]{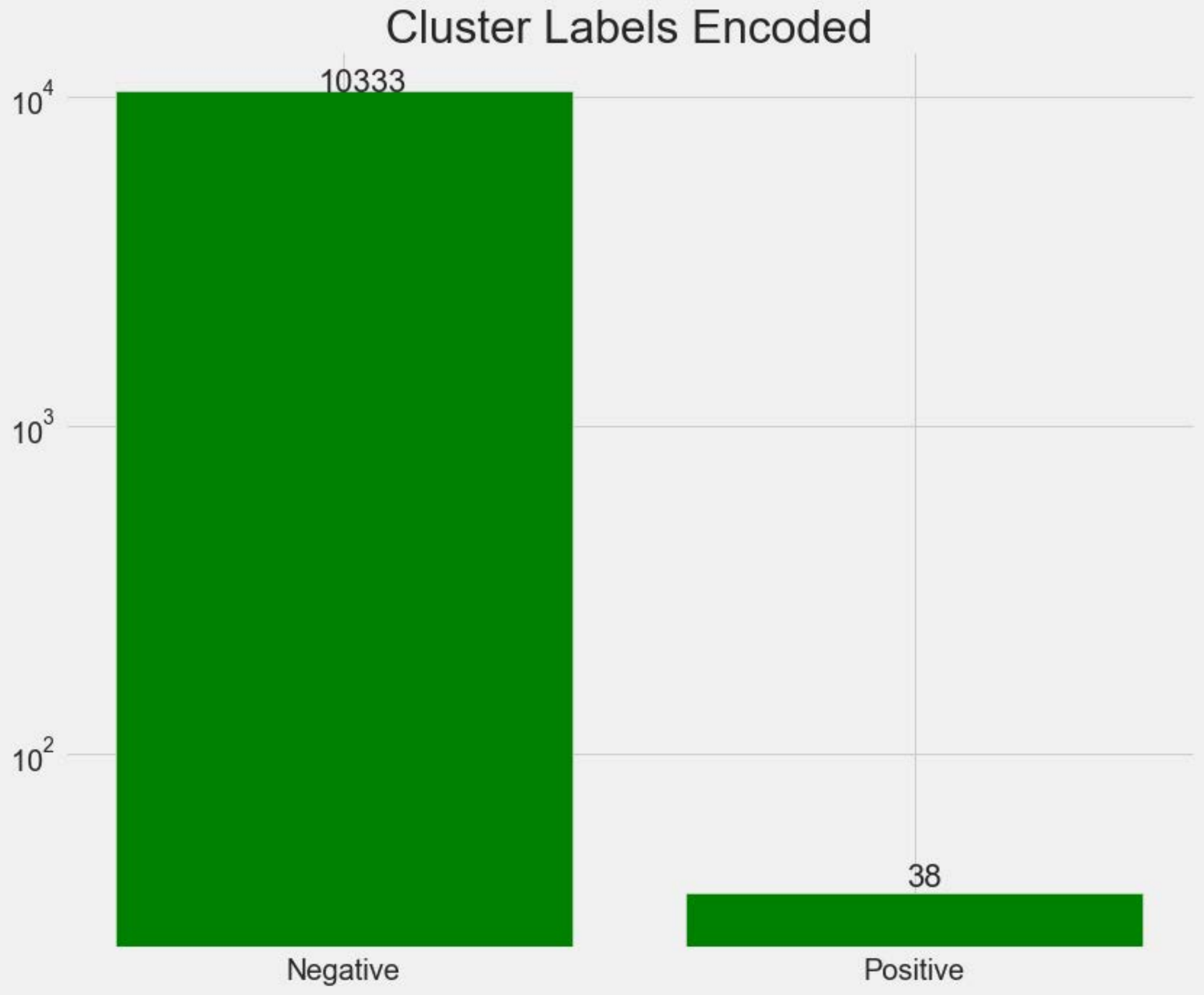}
        \caption{Encoded cluster labels for classification task - Castellfollit case study.}
    \end{subfigure}
    \caption{Target labels summary. Y-axis is in log-scale for comparison purposes. The actual values are included as text above each bar.}
    \label{fig:cluster_labels}
\end{figure}

The normalized clustered data of point clouds are used for the training of our machine learning models. Then, a vector ($v$) is constructed that models the target label naively as follows. If the cluster is considered a rockfall event, which can also be denoted as rockfall candidate event, then we set a positive label ($v_i = 1$), and if not we assign a negative label ($v_i = 0$). A summary of cluster labels is depicted in \figurename~\ref{fig:cluster_labels}, where sub-figures (a) and (c) show the distribution of cluster labels before and figures (b) and (d) after encoding in the Degotalls and Castellfollit case studies respectively.

Please note that the rockfall candidates belong in the class where $v_i = 1$ and the various "Limit\_effect", "Unknow", "Vegetation", "Precursor" cases belong in the class where $v_i = 0$.

\subsection{Resampling Techniques}
\label{sec:resamplers}

Rockfall events are characterized by their rarity and thus our task is considered to be an imbalanced classification task. This is also justified by Fig~\ref{fig:cluster_labels}, where the candidate clusters for rockfall represent roughly 1\% and 0.4\% percent of the total cases in Degotalls and Castellfollit respectively. There are two well known approaches to balance the data instances, namely \textit{undersampling} and \textit{oversampling} methods~(\cite{Hernandez2013AnDatasets}). Undersampling removes elements from the majority class, while oversampling creates synthetic samples from the minority class. There are also ensemble methods that undersample the majority class and oversample the minority one simultaneously~(\cite{Junsomboon2017CombiningDataset}). Below, we briefly describe each one of the resamplers examined in this research. Finally, we conclude by utilizing the three best approaches in our subsequent analysis phases, according to their performance.

\begin{description}

\item \textbf{Undersampling Methods}

Naively, the simplest technique to perform undersampling is by doing it randomly, i.e. by removing individual samples belonging to the majority class without any heuristic extraction. More elaborate techniques use some kind of heuristic processing in order to remove elements that are informationally irrelevant to the whole dataset. Some of the aforementioned methods, aggregate distinct samples to undersample the data using well-known clustering techniques, such as (\cite{Yen2009Cluster-basedDistributions}).

In this study, we utilize two undersampling methodologies based on the K-means clustering algorithm. Specifically, we use two of the methods proposed by \cite{Yen2009Cluster-basedDistributions} utilizing the K-means with the number of clusters equal to the number of samples belonging to minority class. The first methodology generates centroids based on K-means clustering and undersamples the majority class by replacing a cluster of majority samples with the cluster centroid of the K-Means algorithm (denoted Cluster Centroids in this writing). The second one generates centers and undersamples the majority class by replacing a cluster of majority samples with the cluster center of the K-Means algorithm (denoted Cluster Representatives).

\item \textbf{Oversampling Methods}

Oversampling techniques perform the balancing of the dataset by focusing on the minority classes. Various oversampling approaches are described in the literature (\cite{Kovacs2019AnDatasets}). One of the mostly utilized oversampling techniques, namely SMOTE, generates synthetic samples by combining features of the nearest neighbours of the minority class~(\cite{Chawla2002SMOTE:Technique}). Numerous variants of the aforementioned algorithm appear in the literature (\cite{Fernandez2018SMOTEAnniversary}). \cite{Kovacs2019AnDatasets} presented a detailed empirical comparison of multiple variants of minority oversampling techniques involving a wide assortment of imbalanced datasets for evaluation purposes. In this study, we focus on some of them, specifically the ones that we utilized for the purpose of this research.

The ADASYN oversampler uses a weighted density distribution $r_x$ for different minority class examples according to their level of difficulty in learning, thereby deciding the number of synthetic samples to be generated for a particular point (\cite{He2008ADASYN:Learning}). In contrast, SMOTE uses a uniform weight for all minority points. The major difference between SMOTE and ADASYN is the difference in the generation of synthetic sample points for minority data points. 

\cite{Barua2013ProWSyn:Learning} proposed a proximity weighted synthetic oversampling technique (ProWSyn) that generates effective weight values for the minority data samples based on the sample’s proximity information. Proximity, in this case, is defined as the distance between the sample and the boundary (\cite{Barua2013ProWSyn:Learning}). Another extension of SMOTE through a new component, an iterative ensemble-based noise filter called Iterative-Partitioning Filter (IPF), is named SMOTE-IPF. SMOTE-IPF is considered to overcome the problems produced by noisy and borderline examples in imbalanced datasets as presented and explained in (\cite{Saez2015SMOTE-IPF:Filtering}).

SMOTE algorithm ignores data distribution and density information which is significant in order to synthesize minority examples correctly. The SMOBD algorithm tackles the aforementioned issues and introduces the effective elimination of the influence of noise (\cite{Cao2011ApplyingLearning}). In addition, the Assembled-SMOTE method implements oversampling by taking into consideration data distribution information in order to avoid the occurrence of overlapping between classes (\cite{Zhou2013AClassification}). 

Other methods, such as the Lee algorithm, generate synthetic samples and decide whether to reject or accept a generated sample by considering their location, i.e. the nearest neighbors of a synthetic sample (\cite{Lee2015AnLearning}). \cite{Batista2004AData} proposed an oversampling method call SMOTE-TomekLinks, which is basically a SMOTE algorithm with the additional application of Tomek links to the oversampled training set as a data cleaning technique. Thus, the aforementioned technique improves the oversampling accuracy by removing examples from both classes. Tomek links are introduced in (\cite{Tomek1976TwoCNN}) and can be defined as follows:

\begin{quote}
Given two examples $E_i$ and $E_j$ belonging to different
classes, and $d(E_i, E_j)$ is the distance between $E_i$ and $E_j$. A $(E_i, E_j)$ pair is called a Tomek link if there is not an example $E_l$, such that $d(E_i, E_l) < d(E_i, E_j)$ or
$d(E_j , E_l) < d(E_i, E_j)$.
\end{quote} 

Furthermore, the CCR algorithm combines cleaning the decision border around minority objects with guided synthetic oversampling to properly detect the minority class examples (\cite{Koziarski2017CCR:Classification}). Moreover, the G-SMOTE method introduced hybrid oversampling, by the notion of doing partially sampling guided by the hidden patterns obtained from minority class and randomization. Highly skewed data distributions are handled by simultaneous over and undersampling (\cite{Sandhan2014HandlingRecognition}). LVQ-SMOTE uses a combination of the SMOTE ovesampler and the feature codebooks obtained by the learning vector quantization in order to generate synthetic samples which occupy more feature space than the other SMOTE variants (\cite{Nakamura2013LVQ-SMOTEData}). Also, a comparable performance can be obtained using the Polynom-fit SMOTE, which oversamples the minority class using polynomial fitting functions (\cite{Gazzah2008NewSets}).

Moreover, the research study of (\cite{Stefanowski2008SelectivePerformance}) proposed a method for selective pre-processing, called SPIDER that combines filtering and oversampling of imbalanced data. Also, SWIM algorithm generates synthetic samples that have the same Mahalanobis distance from the majority class as the currently known minority data samples (\cite{Sharma2018SyntheticImbalance}).

\end{description}

In this study, we examine and experiment with the sixteen resamplers explained above, namely Cluster Centroids, Cluster Representatives, SMOTE, ProWSyn, SMOTE-IPF, SWIM, SMOBD, Lee, ADASYN, Assembled-SMOTE, SMOTE-TomekLinks, CCR, G-SMOTE, LVQ-SMOTE, Polynom-fit-SMOTE and SPIDER.

\subsection{Models}
\label{sec:proposed_models}

This section describes the machine learning models trained and evaluated in the rockfall detection task. In addition, we provide basic information regarding the models utilized. However, a complete explanation and analysis of the algorithmic aspects of each model is out of the scope of this paper. In the following, there is a list containing all the models examined including their hyper-parameters. For clarification purposes, we denote in parenthesis the abbreviation for each model name to be used in this analysis.

\begin{description}
\item \textbf{Linear Discriminant Analysis (LDA)} is a linear decision boundary classifier, whose boundary is generated by fitting class conditional densities to the input data. This particular model fits a Gaussian density to each class. All classes are assumed to share the same covariance matrix~(\cite{Hastie2009TheLearning}). We experiment with solvers such as Singular Value Decomposition (SVD), Least Squares Solution (LSQR) and Eigenvalue Decomposition (EIGEN), and with automatic or no shrinkage parameters. 
    
\item \textbf{Quadratic Discriminant Analysis (QDA)} is a classifier that classifies objects based on a quadratic decision boundary that is generated by fitting class conditional densities to the input data. This particular model, like the Linear Discriminant Analysis, fits a Gaussian density to each class. All classes are assumed to share the same covariance matrix~(\cite{Hastie2009TheLearning}). We utilize the default configuration of this classifier. 

\item \textbf{K-Nearest Neighbors Classifier (KNN)} is a memory-based algorithm. Naively, given a query point $x_0$, we find the $k$ training points $x_{(r)}
,r = 1,... ,k$ closest in distance to $x_0$, and then classify it using a majority vote policy among the $k$ neighbors~(\cite{Hastie2009TheLearning}). We experiment using various numbers of nearest neighbors, such as 1, 3, 5 and 9 neighbors. 

\item \textbf{Gaussian Naive Bayes (GNB)} assumes that given a class $G = j$, the
features $X_k$ are independent: $f_j(X) = \prod_{k=1}^{p} f_{jk}(X_k)$~(\cite{Hastie2009TheLearning}). Gaussian Naive Bayes supports continuous valued features and models each as conforming to a Gaussian (normal) distribution. We utilize the default parameters for this specific classifier.

\item \textbf{Decision Tree Classifier (DT)} partitions the feature space into a set of rectangles, and then fits a simple model on each one. Tree-based classifiers are conceptually simple yet powerful~(\cite{Hastie2009TheLearning}). For this particular classifier, we experiment with two strategies for splitting each node, namely the best split and random split. Regarding the criterion for splitting or alternatively the function to measure the quality of a split, we experiment with Gini impurity (\cite{DAmbrosio2011ConditionalMeasure}) and information gain metrics.

\item \textbf{AdaBoost Classifier (AdaB)} is a meta-estimator that initially fits a classifier on the original input data and then fits additional copies of the classifier on the same data but conditionally. It fits the additional copies where the weights of incorrectly classified instances are adjusted such that subsequent classifiers focus more on difficult cases~(\cite{Freund1997ABoosting, Zhu2009Multi-classAdaBoost}). Multiple numbers of estimators are investigated, such as 10, 50, 100 and 500. The base estimator is chosen to be a simple decision tree classifier with a max tree depth of 5. 
        
\item \textbf{Random Forest Classifier (RF)} is a meta estimator that fits multiple decision tree classifiers on various sub-samples of the original input data and then uses averaging methods to improve the predictive accuracy and control over-fitting issue~(\cite{Hastie2009TheLearning}). Multiple numbers of trees are investigated, such as 10, 100, 500, 1000. Regarding the function to measure the quality of a split of a tree-node, we investigate the Gini impurity and the information gain.

\item \textbf{Support Vector Classifier (SVC)} algorithm constructs a hyperplane or set of hyperplanes in a high-dimensional space, which can be used for classification. Support vectors define the margins of the hyperplanes and are found after an optimization process involving an objective function regularized by an error term and a constraint. The decision boundaries used by Support Vector Machines (SVMs) can be linear or non-linear using kernel functions. Different kernel functions can be specified for the decision function~(\cite{Awad2015SupportClassification}). In this particular case, we experiment with common kernels such as polynomial with multiple distinct degree values ranging from 2 to 8, a radial-basis function, and sigmoid. Regarding the regularization parameter, we have used multiple values, such as 0.1, 1, 10 and 100.
        
\item \textbf{Extra Trees Classifier (ET)} is a meta estimator that fits multiple randomized decision trees on various sub-samples of the original input data and uses averaging methods to improve predictive accuracy and control over-fitting issue~(\cite{Geurts2006ExtremelyTrees}). Different numbers of estimators are investigated, namely 10, 100, 500 and 1000. Also, the function to measure the quality of a split of a tree-node is set to be the Gini impurity (\cite{DAmbrosio2011ConditionalMeasure}) or the information gain.

\item \textbf{XGBoost Classifier (XGB)} is a scalable end-to-end ensemble machine learning framework. It is actually an implementation of gradient boosted decision trees designed in such way as to achieve state-of-the-art results while remaining fast and efficient~(\cite{Chen2016XGBoost:System}). Different values of boosting learning rate are examined, ranging from $10^{-1}$ to $10^{-4}$. In addition, various numbers of gradient boosted trees are used such as 10, 50 and 100, together with a variety of boosters such as \textit{gbtree}, which is a version of a regression tree as a weak learner, \textit{gblinear}, which uses generalized linear regression with L1 and L2 shrinkage, and \textit{dart}, which drops trees in such way as to reduce the over-fitting issue.

\item \textbf{Multi-Layer Perceptron (MLP) Classifier} is a class of feed-forward artificial neural networks~(\cite{Murtagh1991MultilayerRegression}). Alternatively, the MLP classifier can be considered to be a shallow deep neural network. The investigated hyper-parameters are hidden layer sizes of 50, 100, 150 and 200 and activation functions for the hidden layers, namely rectified linear unit (ReLu) $f(x) = max(0,x)$, hyperbolic tangent (TanH) $f(x) = tanh(x)$, and logistic sigmoid $f(x) = \frac{1}{1+exp(-x)}$. Besides, three optimizers are tested namely LBFGS, which is an optimizer of the family of quasi-Newton methods, the stochastic gradient descent (SGD) and Adam, which is a stochastic gradient-based optimizer.

\end{description}

\subsection{Feature Selection}

In this phase of the analysis framework, see Fig \ref{fig:methodology} (d), the best selected and properly parameterized models of each model variant accompanied with each one of the three best resamplers are used as inputs in a grid search algorithm aiming to select the exact number of features that achieve the highest performance. The criterion used for the univariate feature selection is mutual information between two random variables. Mutual information is a non-negative value that measures the dependency between the variables~(\cite{Kraskov2004EstimatingInformation}). According to the aforementioned methodology, we keep only the $k$ highest scoring features, with $k$ ranging from 5 to 37 (all features) in the Degotalls case study and from 5 to 31 (all features) in the Castellfollit case study.

\subsection{Pipeline Selection}
\label{subsec:pipeline}

The input of this stage consists of the multiple machine learning pipelines, each properly parameterized with the best number of features, as shown in Fig \ref{fig:methodology} (e). These pipelines are statistically compared in order to conclude with the best intelligent pipeline to be used in our prototype system. The statistical comparison of the various pipelines is conducted by applying the Friedman, a non-parametric statistical test, and Nemenyi, a post-hoc test as described in (\cite{Demsar2006StatisticalSets}). We run every pipeline $n = 10$ times using 10-fold cross-validation. In each run, we rank the pipelines from $1$ to $n_{models}$. By averaging the ranks table to get the averaged ranks for each pipeline, we can conclude that two models are significantly different if their average ranks differ by at least the critical difference. The critical difference is computed with the following formula:
    
\begin{equation}
    CD = q_{\alpha}\sqrt{\frac{n_{models}*(n_{models} - 1)}{6*n}} ,
\end{equation}

\noindent where $q_{\alpha}$ denotes the critical value of the two tailed Nemenyi test, which depends on the chosen $\alpha$ level of statistical significance and the number of models or in this case the pipelines, $n_{models}$.

\subsection{Feature Importance}
\label{subsec:feature_importance}

In this phase of our framework, see Fig \ref{fig:methodology} (h), we calculate and visualize the feature importance of the input dataset. This step provides significant insights in terms of the interpretability and explainability of the outcome produced by the utilized algorithm. Thus, we utilize a generalized feature importance measurement, the permutation feature importance as introduced and described in (\cite{Breiman2001RandomForests}). The computation of permutation feature importance is done by first calculating a baseline metric, defined by a performance scoring function, and evaluated on a dataset $X$. Next, a feature from the validation set is permuted and the metric is evaluated again. Finally, the permutation importance is defined as the difference between the baseline metric and the metric obtained from permutating the aforementioned feature.

In our study, we utilize an extensive permutation feature importance test in order to get robust results and insights regarding the most significant features utilized by the classifier. Specifically, we use the most statistical significant classifier as concluded from the computations in the \nameref{subsec:pipeline} stage of our framework, depicted in Fig \ref{fig:methodology} (e), with a 10-fold cross validation technique on input data and 100 feature permutations in each fold. Then, we average all the values obtained and express them in percentage values for interpretability purposes. In addition, we use K-means clustering algorithm to group the feature importances obtained in order to get a better view of the most important features and their neighbors.

\section{Experimental Analysis} 
\label{sec:exper}

This section presents the experimental analysis of our designed intelligent framework, based on well-known performance metrics. First, we explain the accuracy metric and then we present evaluation tables for each stage of the framework. Additionally, an ablation study is conducted in order to justify the addition of each component to the final prototype implementation.

\subsection{Performance Metrics}
\label{subsec:perf_metr}

Our aim is to design our models for a classification task, in which a false negative is usually more disastrous than a false positive for preliminary rockfall detection. The nature of our problem is imbalanced as discussed in Section \ref{sec:basic_proc}. For this reason, we utilize a special metric called balanced accuracy, as introduced and explained in (\cite{Brodersen2010TheDistribution}). The authors of the aforementioned study define balanced accuracy as a performance metric for imbalanced classification tasks. Balanced accuracy can be naively defined as the average accuracy obtained on either class.

Based on a confusion matrix

\begin{center}
\renewcommand\arraystretch{1.5}
\setlength\tabcolsep{0pt}
\begin{tabular}
{c >{\bfseries}r @{\hspace{0.7em}}c @{\hspace{0.4em}}c @{\hspace{0.7em}}l}
  \multirow{10}{*}{\parbox{1.1cm}{\bfseries\raggedleft actual\\ value}} & 
    & \multicolumn{2}{c}{\bfseries Prediction outcome} & \\
  & & \bfseries p & \bfseries n & \bfseries total \\
  & p$'$ & \MyBox{True}{Positive (TP)} & \MyBox{False}{Negative (FN)} & P$'$ \\[2.4em]
  & n$'$ & \MyBox{False}{Positive (FP)} & \MyBox{True}{Negative (TN)} & N$'$ \\
  & total & P & N &
\end{tabular}
\end{center}

the balanced accuracy is given by:

\begin{equation}
\label{eq:bl_acc}
    Acc_{b} = \frac{1}{2}\left(\frac{TP}{TP+FN} + \frac{TN}{TN+FP}\right).
\end{equation}

\subsection{Experimental Evaluation}
\label{results}

In this section, we experimentally compare the various machine learning models, which we presented and briefly analyzed in Section \ref{sec:proposed_models}. We first present the performance evaluation of the various resamplers in Table \ref{tbl:resamplers_perfo} and then a thorough evaluation of all the models used in the model selection and parameterization and feature selection phases in Table \ref{tbl:models_perfo}. The entire evaluation process is based on the balanced accuracy metric, denoted in equation \ref{eq:bl_acc}, using a stratified 10-fold cross-validation. 

\subsubsection*{Resamplers}

Table~\ref{tbl:resamplers_perfo} displays the average performance of each resampler in the Degotalls (a) and Castellfollit (b) case studies. The depicted values are the average performances of all machine learning models evaluated with a  stratified 10-fold cross validation procedure, without prior parameterization, paired with each resampler. We average the balanced accuracy metric values resampler-wise, in order to select the three best resamplers to proceed to the model selection and parameterization stage. In both case studies, the Cluster Centroids resampler achieves the highest balanced accuracy while being the most robust method. It achieves a $\overline{Acc_{b}}$ of 0.89 with 3.83\% error and 0.82 with 7.78\% error in the Degotalls and Castellfollit case studies respectively. Observing the two sub-tables in Table~\ref{tbl:resamplers_perfo}, the results obtained in both cases do not seem to differ much in terms of ranking.

\begin{table}
    \caption{$Acc_{b}$ metric summary for each resampling method. $\overline{Acc_{b}}$ denotes the average value of $Acc_{b}$ of the 10-fold cross validation. With dark gray we denote the best, with lighter gray the second best and with pale gray the third best.}
    \label{tbl:resamplers_perfo}
    \begin{minipage}[t]{.5\textwidth}
      \caption*{(a) Degotalls case study.}
      \centering
      
      \resizebox{\columnwidth}{!}{%
        \begin{tabular}{|l|r|r|}
        \hline
        \textbf{Resampling Method} &      \textbf{$\overline{Acc_{b}}$} &       \textbf{Error (\%)}\\
        \hline
        Cluster Centroids       &        \cellcolor[gray]{0.55} 0.89 &              \cellcolor[gray]{0.55} 3.83 \\ \hline
        ProWSyn                 &        \cellcolor[gray]{0.7} 0.87 &            \cellcolor[gray]{0.7}  5.34 \\ \hline
        SMOTE-IPF               &        \cellcolor[gray]{0.85} 0.87 &            \cellcolor[gray]{0.85}  5.56 \\ \hline
        SWIM                    &         0.86 &              5.65 \\ \hline
        SMOBD                   &         0.86 &              6.29 \\ \hline
        Lee                     &         0.86 &              5.48 \\ \hline
        ADASYN                  &         0.86 &              6.28 \\ \hline
        Assembled-SMOTE         &         0.86 &              6.16 \\ \hline
        SMOTE                   &         0.86 &              5.85 \\ \hline
        SMOTE-TomekLinks        &         0.86 &              6.56 \\ \hline
        CCR                     &         0.85 &              7.91 \\ \hline
        G-SMOTE                 &         0.85 &              7.15 \\ \hline
        LVQ-SMOTE               &         0.84 &              6.20 \\ \hline
        Polynom-fit-SMOTE       &         0.83 &              7.56 \\ \hline
        SPIDER                  &         0.80 &              6.29 \\ \hline
        Cluster Representatives &         0.78 &              7.67 \\ \hline
        
        \end{tabular}
        }
    \end{minipage}%
    \begin{minipage}[t]{.5\textwidth}
      \centering
        \caption*{(b) Castellfollit case study.}
        
        \resizebox{\columnwidth}{!}{%
        \begin{tabular}{|l|r|r|}
        \hline
        \textbf{Resampling Method} &      \textbf{$\overline{Acc_{b}}$} &       \textbf{Error (\%)}\\
        \hline
        Cluster Centroids       &         \cellcolor[gray]{0.55} 0.82 &             \cellcolor[gray]{0.55} 7.78 \\\hline
        ProWSyn                 &         \cellcolor[gray]{0.7} 0.71 &             \cellcolor[gray]{0.7} 16.94 \\\hline
        CCR                     &         \cellcolor[gray]{0.85} 0.70 &             \cellcolor[gray]{0.85} 15.88 \\\hline
        SWIM                    &         0.70 &             14.90 \\\hline
        Lee                     &         0.68 &             17.52 \\\hline
        SMOTE                   &         0.68 &             18.13 \\\hline
        LVQ-SMOTE               &         0.68 &             18.33 \\\hline
        ADASYN                  &         0.68 &             18.50 \\\hline
        SMOBD                   &         0.68 &             18.86 \\\hline
        SMOTE-IPF               &         0.68 &             17.76 \\\hline
        SMOTE-TomekLinks        &         0.68 &             17.33 \\\hline
        Assembled-SMOTE         &         0.67 &             18.08 \\\hline
        G\_SMOTE                 &         0.66 &             19.83 \\\hline
        Polynom-fit-SMOTE       &         0.65 &             18.00 \\\hline
        SPIDER                  &         0.58 &             17.29 \\\hline
        Cluster Representatives &         0.56 &             18.28 \\\hline

        \end{tabular}
        }
    \end{minipage} 
\end{table}

\subsubsection*{Model Selection}

Table~\ref{tbl:models_perfo} shows the performance evaluation of the model selection and parameterization phase in the Degotalls (a) and Castellfollit (b) case studies. Specifically, it displays only the six best methods for each case study. For clarification purposes, the full tables are included in Tables \ref{tbl:models_perfo_dego_app} and \ref{tbl:models_perfo_castell_app} respectively in the \nameref{S2_Appendix}. For the selection of the best hyper-parameters of each model, a grid search algorithm is utilized using a stratified 10-fold cross-validation procedure for each available combination of parameters. Please note that the depicted values in the above-mentioned table are the balanced accuracy measurements of another stratified 10-fold cross-validation procedure on the dataset utilizing only the best properly configured models that resulted from the aforementioned grid search. It is clear that in the Degotalls case study, the XGBoost classifier parameterized the best, accompanied by the SMOTE-IPF oversampler, noted as XGB-SMOTE\_IPF in Table \ref{tbl:models_perfo} (a), and performed better than all the other models while remaining robust. It achieves a $\overline{Acc_{b}}$ of 0.94 with a 3.83\% error score. Regarding the best hyper-parameters of the XGBoost classifier, a linear booster, namely gblinear, with a learning rate of 0.1 with 50 estimators, was chosen. On the other hand, in the Castellfollit case study, the Linear Discriminant Analysis classifier paired with the ProWSyn oversampler, noted as LDA-ProWSyn in Table \ref{tbl:models_perfo} (b), appears to be the most robust technique in terms of balanced accuracy, achieving a $\overline{Acc_{b}}$ of 0.93 with 0.36\% of error, using Eigenvalue Decomposition as the solver and with an automatic shrinkage parameter.

\subsubsection*{Feature Selection}

In addition, Table \ref{tbl:models_perfo} also displays the performance results of the feature selection phase in the Degotalls (a) and Castellfollit (b) case studies, in which each properly parameterized model accompanied by a resampler is evaluated using the best number of features over a stratified 10-fold cross-validation process. In the first case study, the XGBoost classifier accompanied by the SMOTE-IPF oversampler with 35 features seems to be the best performing model, while in the second best model is the XGBoost classifier paired with ProWSyn with 30 features, as displayed in Table \ref{tbl:models_perfo}.

Generalizing, we could say that in both case studies the top performing methods are approximately the same utilizing around the same percentage of features compared to the total available number of features in each case.

\begin{table}[h!]
\caption{$Acc_{b}$ metric summary. $\overline{Acc_{b}}$ denotes the average value of $Acc_{b}$ of the 10-fold cross validation. With dark gray we denote the best model in terms of accuracy and then robustness and with lighter gray the second best for each phase. Features column display the number of features utilized by each algorithm to achieve this score.} \label{tbl:models_perfo}
\begin{subtable}{\textwidth}

\centering
\caption*{(a) Degotalls case study.}
\resizebox{\textwidth}{!}{%
   \begin{tabular}{| l | c |c | c | c | c |}
       \hline
       \multicolumn{1}{|c|}{\centering \textbf{Method}}  & \multicolumn{2}{|c|}{\textbf{Model Parameterization}}  & 
        \multicolumn{3}{|c|}{\textbf{Feature Selection}}\\ \cline{2-6}  {} & \textbf{$\overline{Acc_{b}}$}&  \textbf{Error (\%)} & \textbf{$\overline{Acc_{b}}$} &  \textbf{Error (\%)} & \textbf{Features}\\
        \hline
      XGB-SMOTE\_IPF         &     \cellcolor[gray]{0.55}0.94 &               \cellcolor[gray]{0.55}3.83 &    \cellcolor[gray]{0.55}0.95 &              \cellcolor[gray]{0.55}3.78 & \cellcolor[gray]{0.55}35 \\\hline
        MLP-SMOTE\_IPF         &     0.94 &               5.54 &    0.95 &              5.64 & 35 \\\hline
        KNN-ClusterCentroids  &     0.94 &               4.29 &    0.95 &              4.09 & 36 \\\hline
        XGB-ProWSyn           &     \cellcolor[gray]{0.7}0.94 &               \cellcolor[gray]{0.7}3.84 &   \cellcolor[gray]{0.7}0.95 &              \cellcolor[gray]{0.7}3.68 & \cellcolor[gray]{0.7}35 \\\hline
        SVC-ProWSyn           &     0.94 &               5.68 &    0.95 &              4.14 & 17 \\\hline
        LDA-SMOTE\_IPF         &     0.94 &               4.13 &    0.95 &              4.07 & 31 \\\hline
       
    \end{tabular}
  }
\end{subtable}

\bigskip
\begin{subtable}{\textwidth}

\centering
\caption*{(b) Castellfollit case study.}
\resizebox{\textwidth}{!}{%
   \begin{tabular}{| l | c |c | c | c | c |}
       \hline
       \multicolumn{1}{|c|}{\centering \textbf{Method}}  & \multicolumn{2}{|c|}{\textbf{Model Parameterization}}  & 
        \multicolumn{3}{|c|}{\textbf{Feature Selection}}\\ \cline{2-6}  {} & \textbf{$\overline{Acc_{b}}$}&  \textbf{Error (\%)} & \textbf{$\overline{Acc_{b}}$} &  \textbf{Error (\%)} & \textbf{Features}\\
        \hline
     XGB-ProWSyn  &     0.93 &               8.11 &     \cellcolor[gray]{0.55}0.94 &               \cellcolor[gray]{0.55}6.92 &         \cellcolor[gray]{0.55}30 \\\hline
LDA-ProWSyn  &     \cellcolor[gray]{0.55}0.93 &               \cellcolor[gray]{0.55}0.36 &    \cellcolor[gray]{0.7}0.93 &              \cellcolor[gray]{0.7}3.98 &        \cellcolor[gray]{0.7}30 \\\hline
MLP-CCR      &     0.93 &               8.21 &    0.93 &              8.18 &        30 \\\hline
LDA-CCR      &     \cellcolor[gray]{0.7}0.93 &               \cellcolor[gray]{0.7}0.54 &    0.93 &             4.06 &        29 \\\hline
XGB-CCR      &     0.92 &               8.11 &    0.92 &              8.17 &        30 \\\hline
SVC-ClusterCentroids  &     0.91 &               8.18 &    0.91 &              8.26 &        30 \\\hline

    \end{tabular}
}
\end{subtable}

\end{table}

\subsubsection*{Pipeline Selection}

\begin{figure}
    \centering
    \begin{subfigure}[t]{0.9\textwidth}
        \centering
         \includegraphics[width = \textwidth]{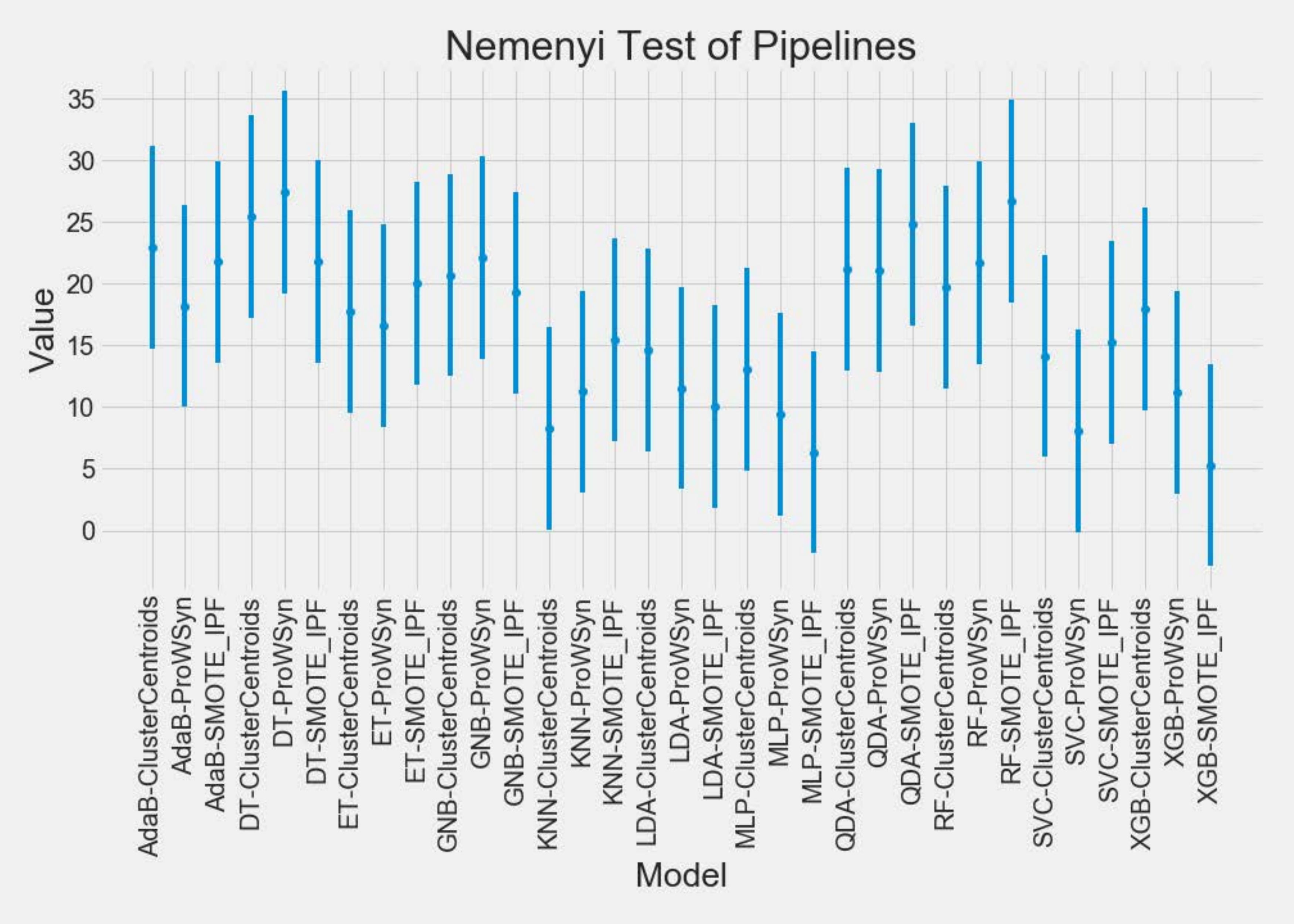}
        \caption{Degotalls case study.}
    \end{subfigure}%
    
    \begin{subfigure}[t]{0.9\textwidth}
        \centering
        \includegraphics[width = \textwidth]{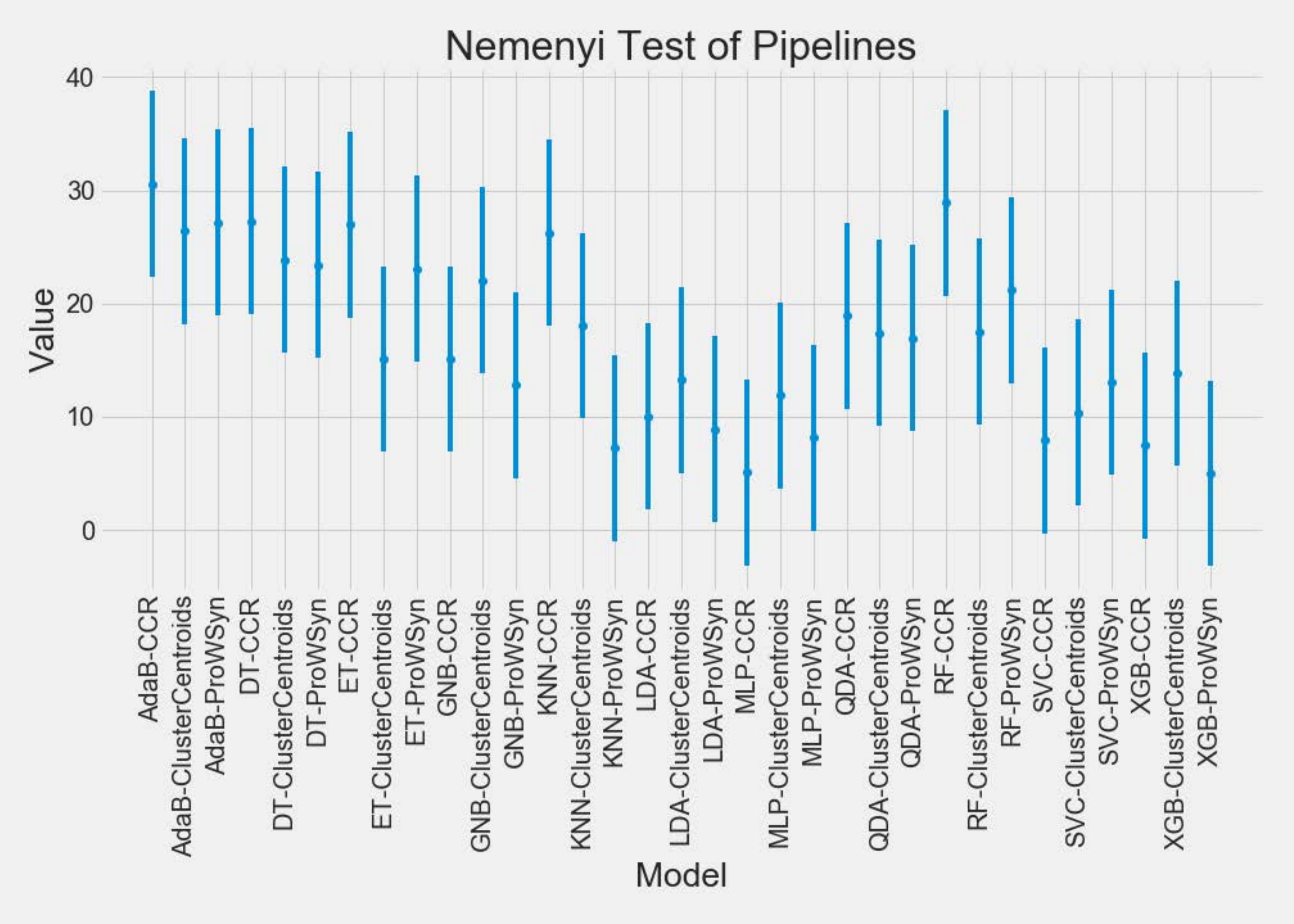}
        \caption{Castellfollit case study.}
    \end{subfigure}
    \caption{Nemenyi test - scores of model pipelines.}
    \label{fig:nemenyi}
\end{figure}

In Fig~\ref{fig:nemenyi}, there is an illustration of the Friedman and Nemenyi techniques, which are computed as described in Section \ref{subsec:pipeline}, for the Degotalls (a) and Castellfollit (b) case studies. This process provides enough evidence to identify which of the methods are more statistically significant than the others. We use a critical distance corresponding to 95\% statistical significance. Please note that between two algorithmic approaches there is a statistical significance when the lines do not overlap. The best algorithm is the one with the minimum rank value. 

Fig \ref{fig:nemenyi} (a) shows that the XGBoost classifier properly parameterized and paired with the SMOTE-IPF oversampler with 35 features performs significantly better than the majority of the machine learning pipelines examined. In addition, the MLP classifier paired with the SMOTE-IPF undersampler with 35 features could also be an acceptable solution.

Regarding the second case study in Fig \ref{fig:nemenyi} (b), the XGBoost classifier properly parameterized and paired with the ProWSyn oversampler with 30 features is the best machine learning pipeline by a significant margin. Furthermore, the MLP classifier accompanied by the CCR oversampler with 30 features could also be an alternative acceptable solution.

\subsubsection*{Feature Importance}

\begin{figure}
    \centering
    \begin{subfigure}[t]{0.9\textwidth}
        \centering
        \includegraphics[width = \textwidth]{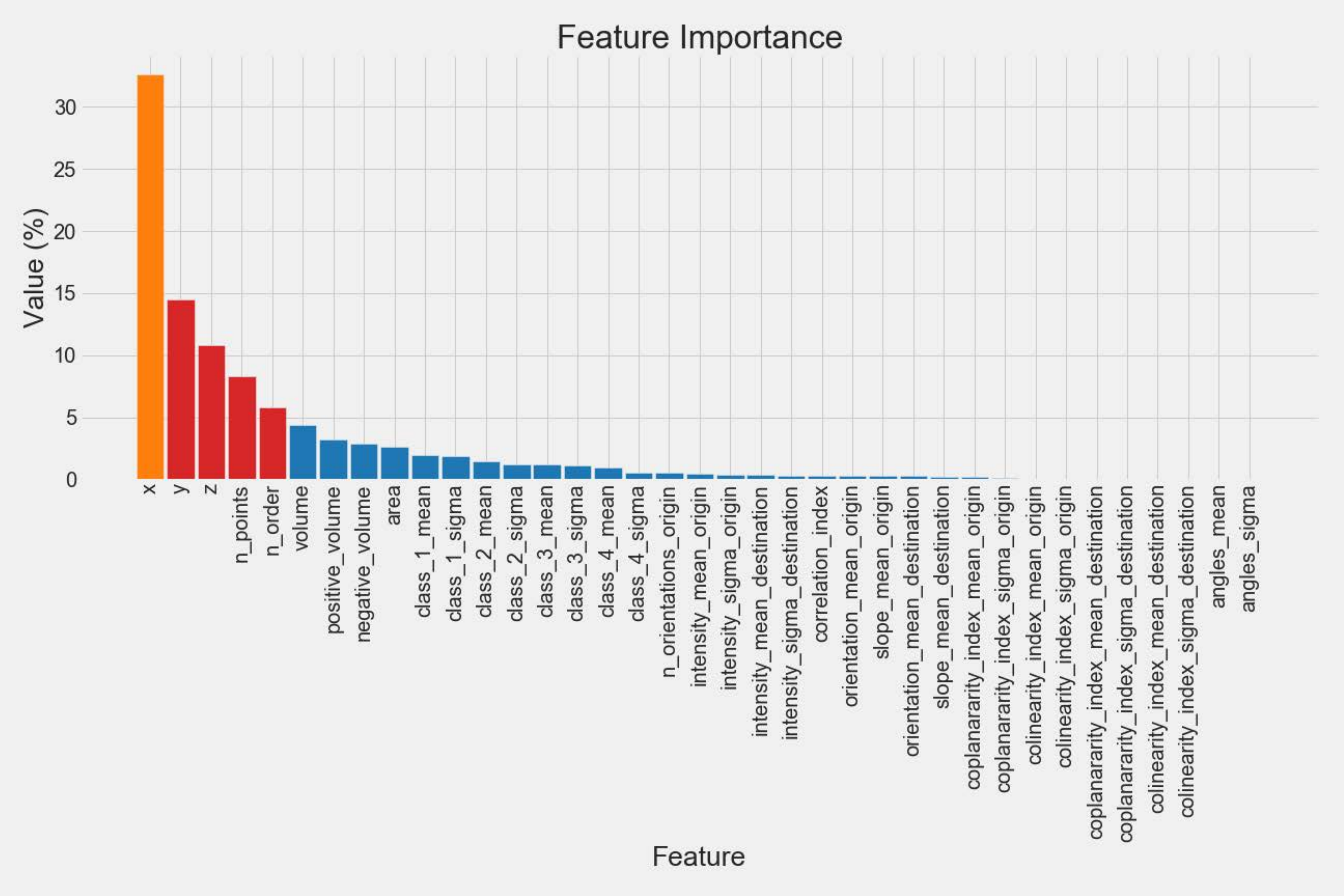}
        \caption{Degotalls case study.}
    \end{subfigure}%
    
    \begin{subfigure}[t]{0.9\textwidth}
        \centering
        \includegraphics[width = \textwidth]{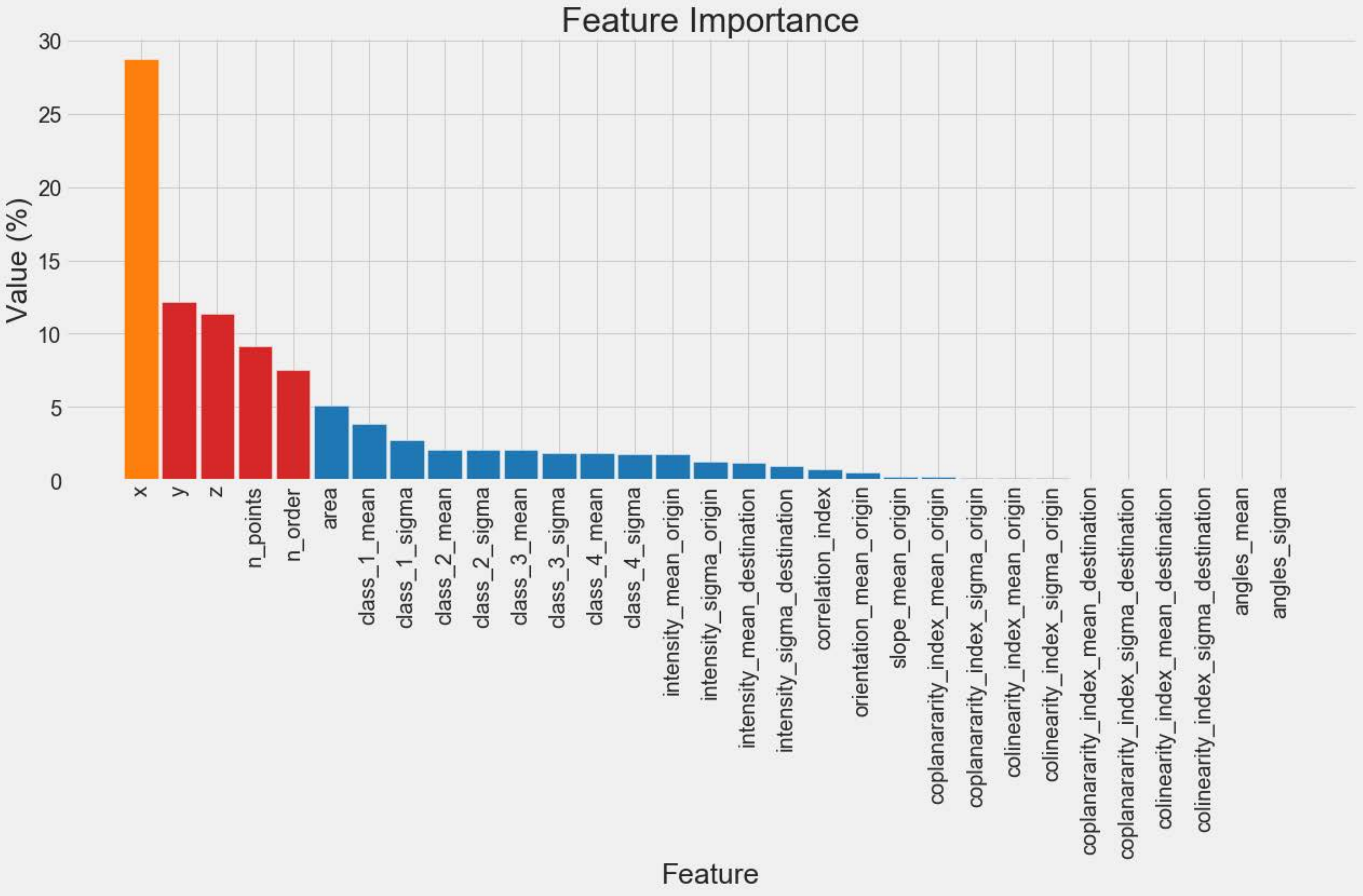}
        \caption{Castellfollit case study.}
    \end{subfigure}
    \caption{Feature Importance. Orange color denotes high importance, blue and red denote mid and low importance respectively.}
    \label{fig:feat_importance}
\end{figure}

In Fig~\ref{fig:feat_importance}, there is a visualization of permutation feature importance technique, as described and explained in Section \ref{subsec:feature_importance}. Please note that the colorization refers to the distinct clusters obtained by the K-means algorithm. For the calculation of the importance of feature in the Degotalls case study, we use the XGBoost classifier paired with the SMOTE-IPF oversampler with 35 features, which we concluded performed better than the majority of all other methods tested, while for the Castellfollit case study, we utilize the XGBoost classifier paired with the ProWSyn oversampler with 30 features.

Observing the sub-figures (a, b) in Fig~\ref{fig:feat_importance}, we can conclude that the most important features for each case study appear to be the same. Moreover, the highly important features for rockfall event detection are the coordinates of the point clouds. 

\subsection{Ablation Study}
\label{subsec:ablation}

An ablation study is crucial for the development of an intelligent system, because it helps to understand the contribution of each component to the overall system. Table~\ref{tbl:ablation} portrays how the balanced accuracy metric is increased with the addition of each proposed component in our analysis framework, in the Degotalls (a) and Castellfollit (b) case studies. It shows the increment of the aforementioned metric across all models ($\overline{Acc_{b}}$) and the best performing model ($Acc_{b}^{best}$) in each phase. The baseline is considered a state in which all the models are trained and evaluated without prior resampling of data, parameterization and feature selection. It is clear in both case studies that the addition of each component provides an interesting performance increment, improving the accuracy and robustness of the utilized method.

\begin{table}[h!]
    \caption{Ablation study - results based on $Acc_{b}$ metric.}
    \label{tbl:ablation}
    \begin{minipage}[t]{.5\textwidth}
      \caption*{(a) Degotalls case study.}
      \centering
      
      \resizebox{\columnwidth}{!}{%
        \begin{tabular}{|l|r|r|r|}
\hline
\textbf{Method} &      \textbf{$\overline{Acc_{b}}$} &       \textbf{Error (\%)} & \textbf{$Acc_{b}^{best}$} \\
\hline
        Baseline &  0.78 &  7.65 &  0.84 \\ \hline
        +Resampling &  0.85 &  3.25 &  0.89  \\ \hline
        +Model Parameterization &  0.89 &  4.65 &  0.94  \\ \hline
        +Feature Selection &  0.91 &  4.45 &  0.95 \\ \hline
    \end{tabular}
        }
    \end{minipage}%
    \begin{minipage}[t]{.5\textwidth}
      \centering
        \caption*{(b) Castellfollit case study.}
        
        \resizebox{\columnwidth}{!}{%
        \begin{tabular}{|l|r|r|r|}
\hline
\textbf{Method} &      \textbf{$\overline{Acc_{b}}$} &       \textbf{Error (\%)} & \textbf{$Acc_{b}^{best}$}\\
\hline
        Baseline &  0.56 &  18.04 &  0.80  \\ \hline
        +Resampling &  0.68 &  8.21 & 0.82  \\ \hline
        +Model Parameterization &  0.79 & 15.93 & 0.93\\ \hline
        +Feature Selection & 0.82 & 11.75 & 0.94 \\ \hline
    \end{tabular}
        }
    \end{minipage} 
\end{table}

\subsection{Prototype Implementation}
\label{sec:prototype}

In this section, we present the prototype implementation of a web-based system appropriate for rockfall detection. The final system utilizes the best intelligent pipeline according to our performance evaluation.

\begin{figure}[H]
    \centering
    \includegraphics[width =  0.9\textwidth]{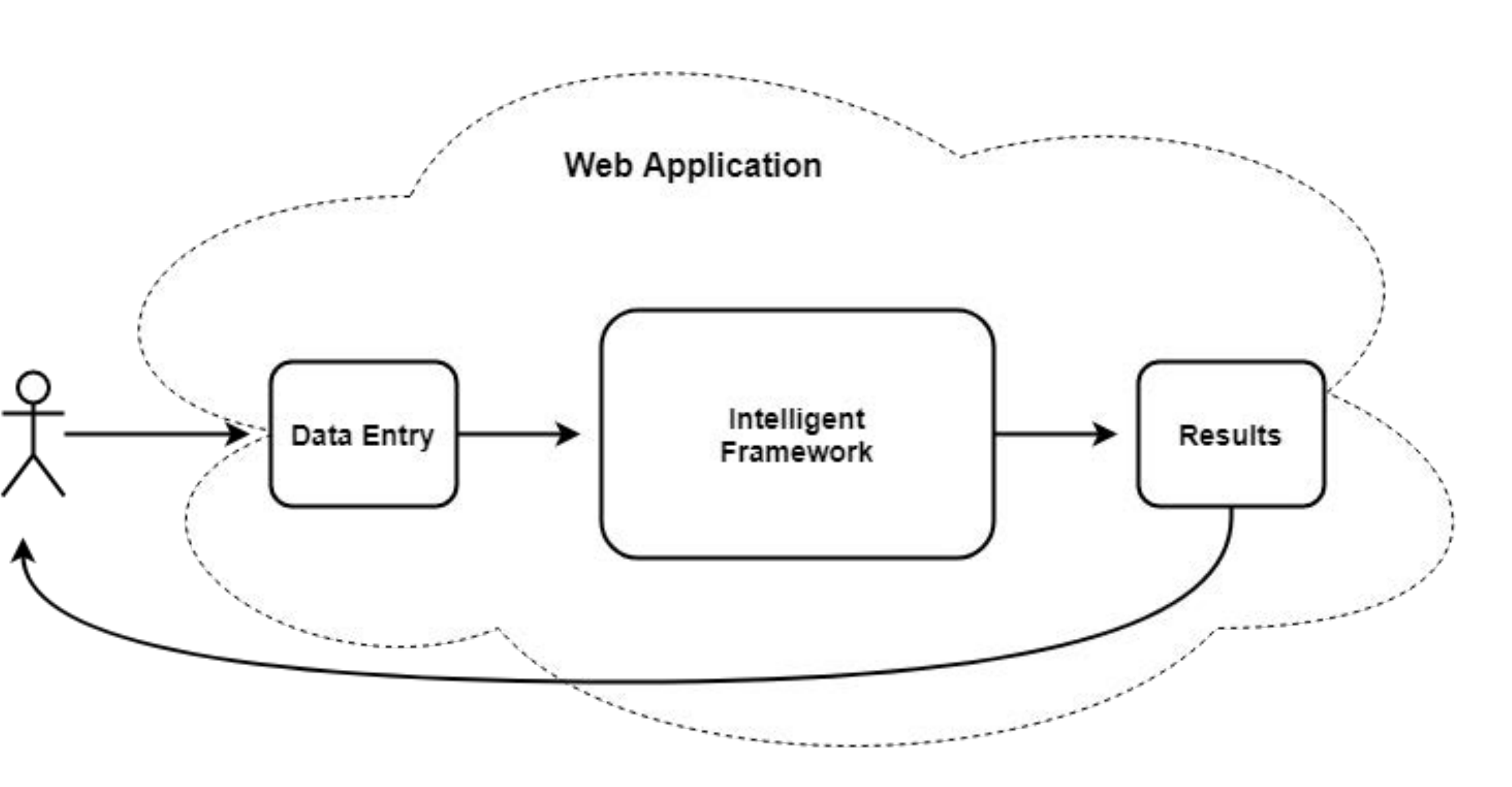}
    \caption{Rockfall Detection - Web Application Architecture.}
    \label{fig:web_app_arch}
\end{figure}

\begin{figure}
    \centering
    \includegraphics[width = 0.9\textwidth]{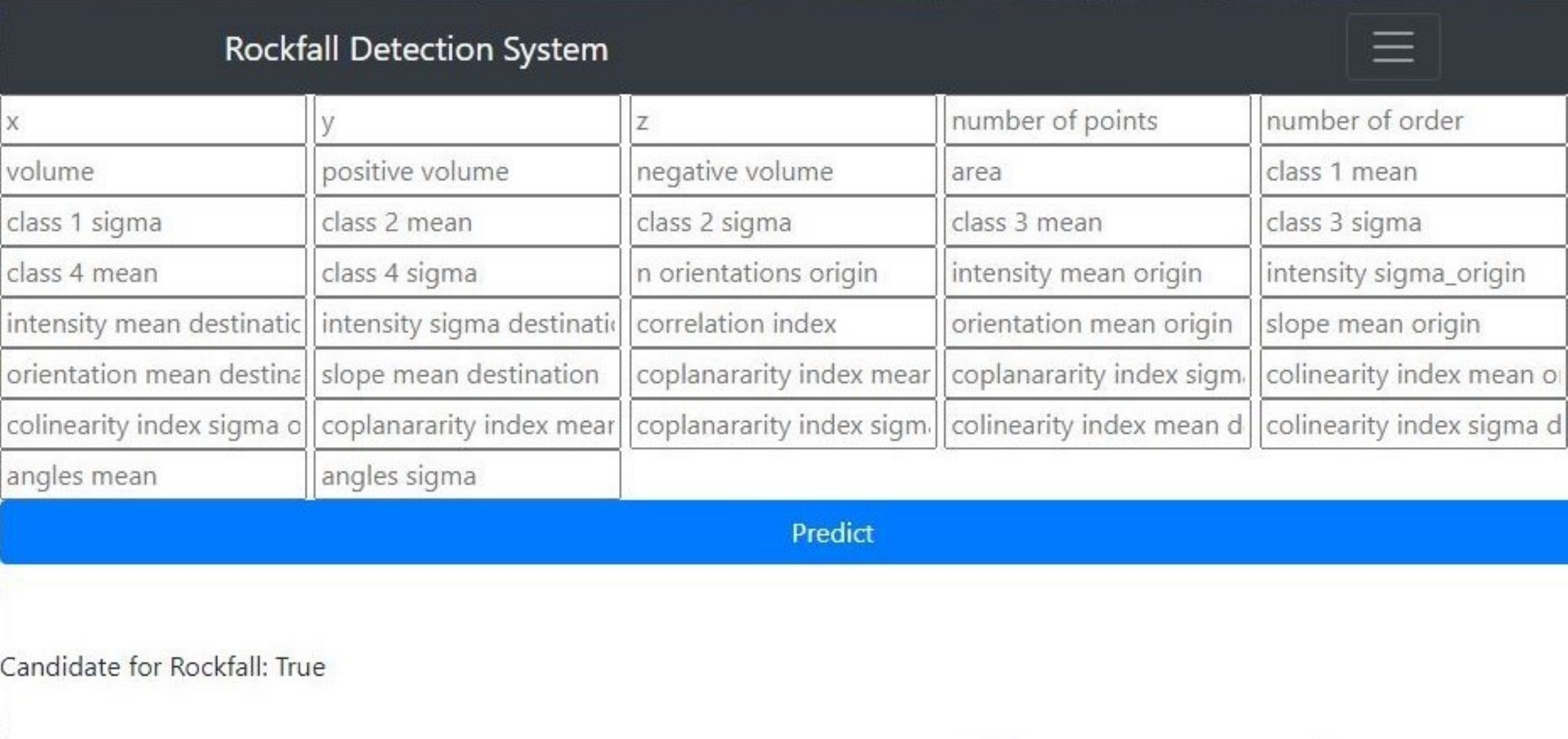}
    \caption{Rockfall Detection - Web Application.}
    \label{fig:web_app}
\end{figure}

In order to demonstrate the effectiveness of our intelligent system, we have developed a simple application that detects whether the input point cloud data are considered candidates for a rockfall event. The architecture of our prototype system for visualizing the predictions can be found in Fig~\ref{fig:web_app_arch}. In the front-end, displayed in Fig \ref{fig:web_app}, there is a simple but properly designed table, which is used to source input feature values and a button that triggers the initialization of the detection procedure. In the back-end, we normalize the data entered by the users, if needed, and generate a prediction using our intelligent framework, which is the pre-trained and loaded machine learning pipeline. Finally, the results are displayed to the user in a fast and accurate manner.

The web application itself can be found at this \href{https://github.com/thzou/rockfall_detection}{GitHub repository}. In addition, we have developed our models and conducted our experiments, utilizing Python programming language. In particular, for building the models, we mainly use the Scikit-learn Python library~(\cite{JMLR:v12:pedregosa11a}). For visualization purposes, we use the Matplotlib library (\cite{Hunter2007}). Regarding the web application, we use Flask (\cite{ThePalletsProjects2010}), a micro web framework.

\section{Final Observations} \label{sec:final_observations} 
In this section, we provide the final observations of our experiments and analysis. In the particular case studies, we conclude that some of the resampling methods and machine learning models examined could be used in order to form intelligent pipelines to detect rockfall events. In the case of the Degotalls mountain wall TLS data, we come to the conclusion that the XGBoost classifier using a linear booster, namely \textit{gblinear}, with a learning rate of 0.1 and 50 estimators utilizing 35 features of the input dataset accompanied by the SMOTE-IPF resampler, achieved the best outcome, with a robust balanced accuracy score of 95\%. Furthermore, in the Castellfollit case study, we conclude that the XGBoost classifier with a linear booster, a learning rate of 0.1 and 100 estimators utilizing 30 features accompanied by the ProWSyn resampler achieved the best balanced accuracy score of 94\%. 

Intuitively, we could say that the second-order gradients and the advanced regularization of the XGBoost algorithm during the learning phase, help to identify better the data relations. In addition, using the \textit{gblinear} booster, the algorithm builds multiple regularized linear models to later incorporate them in a generalized linear model with advanced regularization. It seems that learning inner linear models and then additively producing a generalized linear model facilitates the achievement of a higher accuracy than the other algorithms in both case studies while remaining fast. Additionally, the Castellfollit case study appears to be more difficult than the Degotalls case study in terms of model learning, because of its data imbalance of 0.4\% compared to 1\% in the Degotalls study. This can also be justified by the fact that the XGBoost algorithm utilized twice as many estimators in Castellfollit than Degotalls.

Moreover, the ablation study, portrayed in Table \ref{tbl:ablation} justifies the need for each component of our system. The increment in balanced accuracy metric seems to be significant with the addition of the special framework stages, namely resampling, model selection and parameterization and feature selection. Specifically, the most significant addition appears to be the resampling module of our system, displaying an increment in accuracy from the baseline of approximately 9\% and 21\% in the Degotalls and Castellfollit case studies respectively. Additionally, in general, an ablation study provides an interesting view of how the developed intelligent system behaves by adding additional components. This is a good practice to identify the performance gains of a developed system.

\begin{figure}[H]
    \centering
    \includegraphics[width = \textwidth]{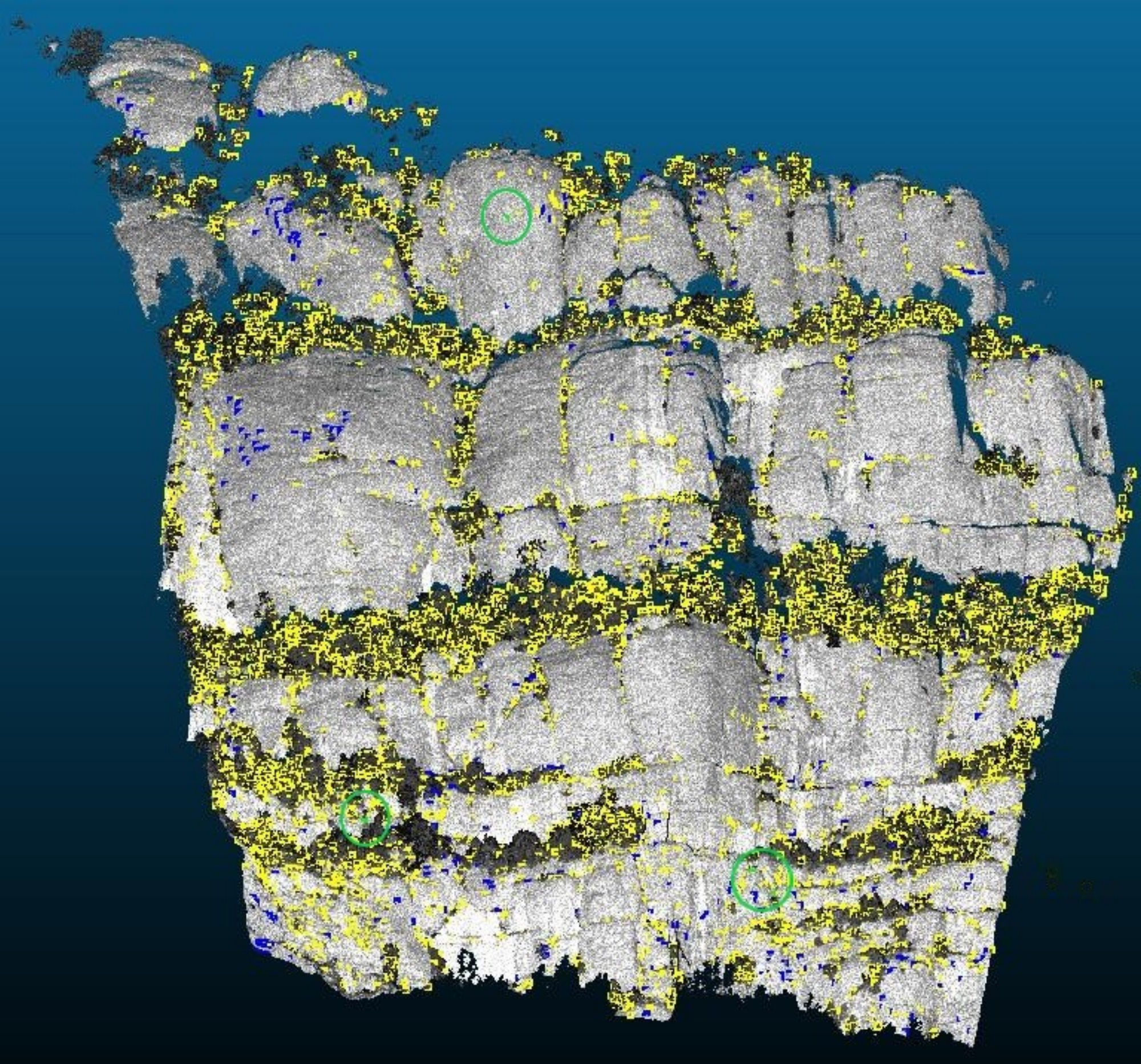}
    \caption{Point Clouds Visualization - Degotalls case study. We denote the correctly and incorrectly classified instances with yellow and blue colors respectively. We further highlight them in green and circle, the events that are initially rockfall events  but were incorrectly classified as not rockfall.}
    \label{fig:degotalls_point_clouds}
\end{figure}

In order to provide further evidence for our developed framework, we visualize the predicted point cloud data along with the original data of the Degotalls case study, as displayed in Fig \ref{fig:degotalls_point_clouds}. We chose to display the former visualization because it includes four incorrectly classified rockfall events representing roughly 6\% of the total rockfall events in the dataset, as depicted in Fig~\ref{fig:cluster_labels} (a). For clarification purposes, in the Castellfollit case study, all rockfall events in the initial dataset were correctly classified by the XGBoost classifier using 30 features paired with the ProWSyn resampler, which is the best selected pipeline, as discussed above.

\section{Conclusion and Future Work} 
\label{sec:synopsis}

Rockfall events are considered to be a great hazard in multiple regions across the world. People working in the geology domain put a lot of effort into identifying such events. However, rockfall event detection utilizing data from remote sensors such as LiDAR or commercial cameras still need specialized analysis tools that can improve the accuracy and efficiency of the distinction of such events from other changes triggered by edge effects, data noise, vegetation, etc. We believe that an intelligent decision assisting tool would facilitate geologists work. However, the development of such a tool seems to be a challenging task, due to the inherent nature of the data.

In this research, an extensive analysis of several algorithmic approaches to the detection of the rockfall events has been presented. In particular, we focus on the development of an end-to-end machine learning framework able to analyze clustered point cloud data from a TLS device. In the presented framework, due to the imbalanced nature of the rockfall event detection, we implement various resampling methodologies, parameterize several machine learning models and create intelligent pipelines to tackle the abovementioned task.

We experimentally evaluate our framework on two case studies, involving data from TLS measurements of a cliff at the Degotalls, Montserrat Massif in Barcelona and the cliff at the Castellfollit de la Roca in Girona, both located in Spain. We analytically present and analyze our design and experimentally evaluate our approach. These efforts elucidate the elements of the whole procedure needed to develop the backbone of an effective, machine learning based, rockfall detection system. We reveal the possibility of developing a practical system which is useful for rockfall detection.

Our study offers the opportunity to elucidate certain important issues and provide specific scientific and technological solutions at the intersection of geology and machine learning. There is significant potential for further additional research regarding the identification of rockfalls. Additionally, raw point cloud data carry an enormous amount of information and utilizing such information effectively and efficiently remains a challenging task. For this, a system utilizing raw point cloud data as input is under construction, in which the proposed methodology is used combined with various added machine learning and neural network based methodologies in order to detect rockfall events directly from temporal point cloud data, showing good potential. Analyzing raw point cloud data without any grouping techniques using sophisticated deep learning architectures may bring advances in rockfall detection, because of the complete utilization of the inner topological features of the data.

\section*{Acknowledgements}

This project has received funding from the European Union’s Horizon 2020 research and innovation programme under the Marie Skłodowska-Curie grant agreement No 860843. In addition, the authors want to acknowledge the support from PROMONTEC (CGL2017-84720-R AEI/FEDER, UE) and SALTEC CGL2017-85532-P (AEI/FEDER, UE) projects, founded by the Spanish MINEICO, and  AGAUR project 2016 DI 069 (Agència de Gestió d’Ajuts Universitaris i de Recerca). Data from Castellfollit de la Roca were acquired with the support of the Spanish Ministry of Science and Education (pre-doctoral grant 2004-1852) and funded by the Natural Park of the Garrotxa Volcanic Field (PNZVG) and the following projects: MEC CGL2006-06596 (DALMASA), TopoIberia CSD2006-0004/Consolider-Ingenio2010, MEC CGL2010-18609 (NUTESA). Data from Montserrat Massif were funded by the Institut Cartogràfic i Geològic de Catalunya (ICGC). Anna Puig and Maria Salamó also acknowledge Generalitat de Catalunya, for its support under project 2017-SGR-341.

\begin{figure}[H]
    \centering
    \includegraphics[scale = 0.6]{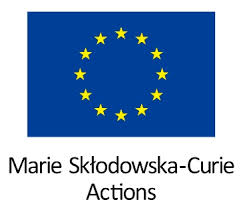}
\end{figure}

\section*{Supporting Information}

\paragraph*{S1 Appendix}
\label{S1_Appendix}

For clarification purposes, we present the full names of the abbreviations used for the data features considered in our study.

\begin{description}

    \item \textbf{x, y, z}: The coordinates of the cluster in Cartesian format.
    \item \textbf{n\_points}: Number of points in the cluster.
    \item \textbf{n\_order}: Number of the order of the cluster.
    \item \textbf{volume}: Total volume of the cluster.
    \item \textbf{positive\_volume}: Positive volume of the cluster, volume of material moving away from the observer (backward motion).
    \item \textbf{negative\_volume}: Negative volume, volume of material displaced toward the observer (advance motion).
    \item \textbf{area}: The area of the surface orthogonal to the TLS.
    \item \textbf{class\_n\_mean}: Mean of the points classified as $n$. 0 is noise, 1 is advance and 2 is waste of volume (associated to rockfall).
    \item \textbf{class\_n\_sigma}: Standard deviation of the points classified as $n$. 0 is noise, 1 is advance and 2 is waste of volume (associated to rockfall).
    \item \textbf{class\_all}: Mean of the points classified as any of the classes.
    \item \textbf{n\_orientations\_o}: Number of different orientations in the cluster at time $t$.
    \item \textbf{Intensity\_mean\_o}: TLS intensity reflection mean at time $t$.
    \item \textbf{Intensity\_sigma\_o}: TLS intensity reflection standard deviation at time $t$.
    \item \textbf{Intensity\_mean\_d}: TLS intensity reflection mean at time $t + 1$.
    \item \textbf{Intensity\_sigma\_d}: TLS intensity reflection standard deviation at time $t + 1$.
    \item \textbf{correlation\_i}: Correlation index between difference and distance between points.
    \item \textbf{orientation\_mean\_o}: Orientation dispersion of reference points.
   \item \textbf{slope\_mean\_o}: Average slope of the reference points.
    \item \textbf{orientation\_mean\_d}: Orientation dispersion of compared points.
    \item \textbf{slope\_mean\_d}: Average slope of the compared points.
    \item \textbf{coplanararity\_i\_mean\_o}: Average of the coplanarity indices of the reference points.
    \item \textbf{coplanararity\_i\_sigma\_o}: Variance of coplanarity of reference points.
    \item \textbf{colinearity\_i\_mean\_o}: Average of the collinearity Indices of the reference points.
    \item \textbf{colinearity\_i\_sigma\_o}: Variance of collinearity of reference points.
    \item \textbf{coplanararity\_i\_mean\_d}: Average of the coplanarity indices of the compared points.
    \item \textbf{coplanararity\_i\_sigma\_d}: Variance of coplanarity with comparative points.
    \item \textbf{colinearity\_i\_mean\_d}: Average of the collinearity indices of the compared points.
    \item \textbf{colinearity\_i\_sigma\_d}: Variance of collinearity of compared points.
    \item \textbf{angles\_mean}: Mean of the angles between reference and compared point.
    \item \textbf{angles\_sigma}: Variance of angles between reference and comparative points.

\end{description}

\normalsize

\paragraph*{S2 Appendix}
\label{S2_Appendix}

In this section, we present the long tables including all the statistics and results, considered in our study.

\begin{table}
\centering
\resizebox{\textwidth}{!}{%
\begin{tabular}{|l|r|r|r|r|r|r|r|}
\hline
\textbf{Feature} &      \textbf{mean} &      \textbf{std} &     \textbf{min} &      \textbf{25\%} &       \textbf{50\%} &       \textbf{75\%} &       \textbf{max} \\
\hline
x                                     &     13.08 &    37.21 &  -56.00 &   -20.10 &     12.52 &     45.39 &    141.47 \\\hline
y                                     &    229.02 &    28.26 &  170.07 &   205.60 &    225.09 &    250.26 &    320.02 \\\hline
z                                     &     32.89 &    31.10 &  -24.31 &     6.64 &     31.21 &     54.33 &    110.87 \\\hline
n\_points                              &  16246.50 &  9036.31 &   16.00 &  8497.00 &  16302.50 &  24167.50 &  31700.00 \\\hline
n\_order                               &     52.23 &   153.54 &   10.00 &    13.00 &     20.00 &     40.00 &   4702.00 \\\hline
volume                                &      0.08 &     0.27 &    0.00 &     0.01 &      0.02 &      0.06 &      9.05 \\\hline
positive\_volume                       &      0.03 &     0.14 &    0.00 &     0.00 &      0.01 &      0.02 &      5.18 \\\hline
negative\_volume                       &      0.05 &     0.16 &    0.00 &     0.00 &      0.01 &      0.04 &      3.86 \\\hline
area                                  &      0.13 &     0.38 &    0.00 &     0.02 &      0.05 &      0.12 &     13.37 \\\hline
class\_1\_mean                          &      1.40 &     0.28 &    1.00 &     1.17 &      1.36 &      1.60 &      2.00 \\\hline
class\_1\_sigma                         &      0.37 &     0.16 &    0.00 &     0.30 &      0.43 &      0.49 &      0.50 \\\hline
class\_2\_mean                          &     25.65 &    21.41 &    0.00 &     8.92 &     19.25 &     39.08 &     88.60 \\\hline
class\_2\_sigma                         &     15.72 &     6.77 &    0.00 &    12.05 &     16.42 &     20.42 &     37.15 \\\hline
class\_3\_mean                          &     45.79 &    27.60 &    0.00 &    23.83 &     44.85 &     66.74 &    100.00 \\\hline
class\_3\_sigma                         &     21.06 &    10.11 &    0.00 &    14.05 &     22.15 &     28.87 &     46.04 \\\hline
class\_4\_mean                          &     28.54 &    23.30 &    0.00 &    10.00 &     23.39 &     41.95 &    100.00 \\\hline
class\_4\_sigma                         &     19.68 &    10.87 &    0.00 &    11.21 &     20.77 &     28.33 &     46.74 \\\hline
n\_orientations\_o                 &     41.54 &    26.04 &    8.02 &    21.44 &     26.66 &     67.40 &    133.70 \\\hline
intensity\_mean\_o                 &    102.09 &    49.64 &   20.94 &    63.27 &     82.51 &    137.15 &    251.53 \\\hline
intensity\_sigma\_o                &     26.85 &    10.77 &    6.28 &    18.51 &     24.63 &     33.61 &     84.52 \\\hline
intensity\_mean\_d            &     99.32 &    50.42 &   21.60 &    58.90 &     79.29 &    138.03 &    254.17 \\\hline
intensity\_sigma\_d           &     25.96 &    11.44 &    1.86 &    17.11 &     24.00 &     33.63 &     94.14 \\\hline
correlation\_i                     &     -0.37 &     0.26 &   -0.97 &    -0.56 &     -0.38 &     -0.19 &      0.69 \\\hline
orientation\_mean\_o               &    171.08 &    84.41 &    0.21 &   119.70 &    179.36 &    216.08 &    359.65 \\\hline
slope\_mean\_o                     &     50.27 &    21.07 &    0.24 &    37.27 &     53.62 &     66.03 &     89.99 \\\hline
orientation\_mean\_d          &    174.20 &    83.88 &    0.01 &   122.05 &    181.33 &    219.56 &    359.82 \\\hline
slope\_mean\_d                &     48.40 &    22.19 &    0.46 &    32.62 &     51.29 &     65.64 &     89.93 \\\hline
coplanararity\_i\_mean\_o       &      1.97 &     0.99 &    0.42 &     1.24 &      1.67 &      2.49 &      6.39 \\\hline
coplanararity\_i\_sigma\_o      &      0.29 &     0.15 &    0.02 &     0.19 &      0.26 &      0.35 &      1.95 \\\hline
colinearity\_i\_mean\_o         &      0.82 &     0.97 &    0.01 &     0.19 &      0.52 &      1.08 &     13.74 \\\hline
colinearity\_i\_sigma\_o        &      0.82 &     1.59 &    0.00 &     0.10 &      0.32 &      0.92 &     40.19 \\\hline
coplanararity\_i\_mean\_d  &      2.10 &     1.01 &    0.51 &     1.35 &      1.79 &      2.62 &      6.28 \\\hline
coplanararity\_i\_sigma\_d &      0.32 &     0.17 &    0.01 &     0.20 &      0.29 &      0.40 &      2.04 \\\hline
colinearity\_i\_mean\_d    &      0.80 &     0.95 &    0.01 &     0.20 &      0.55 &      1.07 &     19.31 \\\hline
colinearity\_i\_sigma\_d   &      0.79 &     1.78 &    0.00 &     0.10 &      0.34 &      0.87 &     54.60 \\\hline
angles\_mean                           &      0.36 &     0.30 &    0.00 &     0.10 &      0.29 &      0.54 &      1.53 \\\hline
angles\_sigma                          &      0.19 &     0.15 &    0.00 &     0.05 &      0.17 &      0.31 &      0.66 \\\hline

\end{tabular}
}
\caption{Summary data statistics - Degotalls case study. The 25\%, 50\% and 75\% denote the 25th, 50th, and 75th percentiles respectively.}
\label{tbl:summary_degotalls}
\end{table} 

\begin{table}
\centering
\resizebox{\textwidth}{!}{%
\begin{tabular}{|l|r|r|r|r|r|r|r|}
\hline
\textbf{Feature} &      \textbf{mean} &      \textbf{std} &     \textbf{min} &      \textbf{25\%} &       \textbf{50\%} &       \textbf{75\%} &       \textbf{max} \\
\hline
x                                     &     61.85 &     53.80 &  -50.94 &     18.06 &     65.13 &    109.43 &     153.41 \\ \hline
y                                     &    181.01 &     29.96 &  116.34 &    155.68 &    196.49 &    205.79 &     230.92 \\ \hline
z                                     &     24.31 &     11.93 &   -3.58 &     14.21 &     23.33 &     34.12 &      52.67 \\ \hline
n\_points                              &  24287.94 &  13713.35 &   14.00 &  12297.50 &  24464.00 &  35866.50 &   47950.00 \\ \hline
n\_order                               &     80.41 &   1673.35 &    7.00 &      8.00 &     12.00 &     24.00 &  138158.00 \\ \hline
area                                  &      0.16 &      0.91 &    0.00 &      0.01 &      0.02 &      0.09 &      38.21 \\ \hline
class\_1\_mean                          &      1.45 &      0.39 &    1.00 &      1.05 &      1.40 &      1.86 &       2.00 \\ \hline
class\_1\_sigma                         &      0.24 &      0.20 &    0.00 &      0.00 &      0.29 &      0.44 &       0.50 \\ \hline
class\_2\_mean                          &     40.06 &     25.75 &    0.00 &     18.64 &     39.43 &     61.00 &      91.86 \\ \hline
class\_2\_sigma                         &     16.15 &      7.97 &    0.00 &     10.43 &     17.18 &     22.44 &      39.50 \\ \hline
class\_3\_mean                          &     32.28 &     28.71 &    0.00 &      4.43 &     26.57 &     54.98 &     100.00 \\ \hline
class\_3\_sigma                         &     14.94 &     11.04 &    0.00 &      5.08 &     14.71 &     23.35 &      46.47 \\ \hline
class\_4\_mean                          &     27.65 &     29.43 &    0.00 &      1.26 &     16.56 &     48.65 &     100.00 \\ \hline
class\_4\_sigma                         &     13.66 &     11.73 &    0.00 &      2.52 &     11.87 &     23.04 &      47.50 \\ \hline
intensity\_mean\_o                 &    118.27 &     46.45 &    0.00 &     85.50 &    115.44 &    146.50 &     255.00 \\ \hline
intensity\_sigma\_o                &     31.69 &     14.75 &    0.00 &     20.66 &     29.90 &     40.71 &     106.77 \\ \hline
intensity\_mean\_d            &     98.77 &     43.72 &    0.00 &     65.38 &     94.00 &    125.44 &     255.00 \\ \hline
intensity\_sigma\_d           &     34.47 &     16.99 &    0.00 &     22.02 &     32.79 &     44.83 &     114.96 \\ \hline
correlation\_i                     &     -0.40 &      0.29 &   -0.97 &     -0.63 &     -0.42 &     -0.20 &       0.73 \\ \hline
orientation\_mean\_o               &    156.17 &     97.21 &    0.07 &     57.08 &    179.20 &    221.43 &     359.97 \\ \hline
slope\_mean\_o                     &     55.38 &     26.81 &    0.16 &     34.86 &     64.08 &     78.22 &      89.88 \\ \hline
coplanararity\_i\_mean\_o       &      2.43 &      0.89 &    0.51 &      1.78 &      2.39 &      2.94 &       7.29 \\ \hline
coplanararity\_i\_sigma\_o      &      0.38 &      0.22 &    0.02 &      0.23 &      0.33 &      0.47 &       1.91 \\ \hline
colinearity\_i\_mean\_o         &      0.52 &      0.82 &    0.01 &      0.09 &      0.22 &      0.63 &      14.93 \\ \hline
colinearity\_i\_sigma\_o        &      0.51 &      1.39 &    0.00 &      0.05 &      0.14 &      0.47 &      46.18 \\ \hline
coplanararity\_i\_mean\_d  &      2.57 &      0.86 &    0.00 &      2.04 &      2.54 &      3.06 &      10.42 \\ \hline
coplanararity\_i\_sigma\_d &      0.57 &      0.47 &    0.00 &      0.26 &      0.44 &      0.74 &       5.05 \\ \hline
colinearity\_i\_mean\_d    &      0.47 &      0.66 &    0.00 &      0.09 &      0.24 &      0.63 &      12.65 \\ \hline
colinearity\_i\_sigma\_d   &      0.49 &      1.04 &    0.00 &      0.05 &      0.16 &      0.54 &      29.23 \\ \hline
angles\_mean                           &      0.43 &      0.36 &    0.00 &      0.11 &      0.32 &      0.73 &       1.54 \\ \hline
angles\_sigma                          &      0.20 &      0.16 &    0.00 &      0.06 &      0.16 &      0.34 &       0.74 \\ \hline

\end{tabular}
}
\caption{Summary data statistics - Castellfollit case study. The 25\%, 50\% and 75\% denote the 25th, 50th, and 75th percentiles respectively.}
\label{tbl:summary_castell}
\end{table} 

\begin{table}

\centering
    \begin{tabular}{| l | c |c | c | c | c |}
       \hline
       \multicolumn{1}{|c|}{\centering \textbf{Method}}  & \multicolumn{2}{|c|}{\textbf{Model Parameterization}}  & 
        \multicolumn{3}{|c|}{\textbf{Feature Selection}}\\ \cline{2-6}  {} & \textbf{$\overline{Acc_{b}}$}&  \textbf{Error (\%)} & \textbf{$\overline{Acc_{b}}$} &  \textbf{Error (\%)} & \textbf{Features}\\
        \hline
      XGB-SMOTE\_IPF         &     0.94 &               3.83 &    0.95 &              3.78 & 35 \\\hline
        MLP-SMOTE\_IPF         &     0.94 &               5.54 &    0.95 &              5.64 & 35 \\\hline
        KNN-ClusterCentroids  &     0.94 &               4.29 &    0.95 &              4.09 & 36 \\\hline
        XGB-ProWSyn           &     0.94 &               3.84 &  0.95 &             3.68 & 35 \\\hline
        SVC-ProWSyn           &     0.94 &               5.68 &    0.95 &              4.14 & 17 \\\hline
        LDA-SMOTE\_IPF         &     0.94 &               4.13 &    0.95 &              4.07 & 31 \\\hline
        MLP-ProWSyn           &     0.94 &               4.07 &    0.95 &              4.09 & 31 \\\hline
        LDA-ProWSyn           &     0.94 &               4.08 &    0.94 &              4.11 & 31 \\\hline
        LDA-ClusterCentroids  &     0.93 &               3.87 &    0.94 &              3.74 & 36 \\\hline
        MLP-ClusterCentroids  &     0.93 &               4.05 &    0.94 &              4.00 & 29 \\\hline
        SVC-ClusterCentroids  &     0.93 &               4.30 &    0.94 &              3.99 & 33 \\\hline
        XGB-ClusterCentroids  &     0.92 &               3.99 &    0.93 &              4.08 & 35 \\\hline
        ET-ClusterCentroids   &     0.92 &               4.01 &    0.93 &              3.94 & 32 \\\hline
        KNN-ProWSyn           &     0.92 &               5.19 &    0.93 &              5.31 & 28 \\\hline
        SVC-SMOTE\_IPF         &     0.92 &               5.26 &    0.93 &              3.94 & 35 \\\hline
        ET-ProWSyn            &     0.90 &               7.44 &    0.91 &              7.31 & 31 \\\hline
        RF-ClusterCentroids   &     0.91 &               3.94 &    0.91 &              4.00 & 33 \\\hline
        KNN-SMOTE\_IPF         &     0.91 &               5.21 &    0.91 &              7.27 & 26 \\\hline
        AdaB-ProWSyn          &     0.89 &               8.29 &    0.90 &              8.20 & 33 \\\hline
        QDA-ClusterCentroids  &     0.89 &               5.56 &    0.90 &              5.52 & 31 \\\hline
        GNB-ClusterCentroids  &     0.88 &               6.07 &    0.90 &              6.76 & 6 \\\hline
        GNB-SMOTE\_IPF         &     0.88 &               7.02 &    0.90 &              5.53 & 6 \\\hline
        GNB-ProWSyn           &     0.87 &               7.84 &    0.89 &              5.56 & 6 \\\hline
        RF-ProWSyn            &     0.86 &               8.67 &    0.89 &              6.68 & 35 \\\hline
        QDA-ProWSyn           &     0.85 &               9.20 &    0.89 &              6.57 & 5 \\\hline
        AdaB-SMOTE\_IPF        &     0.85 &               9.07 &    0.88 &              9.27 & 36 \\\hline
        AdaB-ClusterCentroids &     0.86 &               9.49 &    0.87 &              7.74 & 5 \\\hline
        QDA-SMOTE\_IPF         &     0.85 &               9.23 &    0.87 &              5.90 & 7 \\\hline
        DT-ClusterCentroids   &     0.80 &               6.89 &    0.85 &              6.87 & 30 \\\hline
        ET-SMOTE\_IPF          &     0.84 &               9.01 &    0.84 &              9.04 & 32 \\\hline
        DT-SMOTE\_IPF          &     0.83 &              11.06 &    0.84 &             10.67 & 29 \\\hline
        RF-SMOTE\_IPF          &     0.83 &              10.51 &    0.83 &              9.25 & 23 \\\hline
        DT-ProWSyn            &     0.80 &               9.92 &    0.80 &             10.06 & 32 \\\hline
        
    \end{tabular}

\captionof{table}{$Acc_{b}$ metric summary - Degotalls case study. $\overline{Acc_{b}}$ denotes the average value of $Acc_{b}$ of the 10-fold cross validation. Features column display the number of features utilized by each algorithm to achieve this score.}   \label{tbl:models_perfo_dego_app}
\end{table} 

\begin{table}

\centering
    \begin{tabular}{| l | c |c | c | c | c |}
       \hline
       \multicolumn{1}{|c|}{\centering \textbf{Method}}  & \multicolumn{2}{|c|}{\textbf{Model Parameterization}}  & 
        \multicolumn{3}{|c|}{\textbf{Feature Selection}}\\ \cline{2-6}  {} & \textbf{$\overline{Acc_{b}}$}&  \textbf{Error (\%)} & \textbf{$\overline{Acc_{b}}$} &  \textbf{Error (\%)} & \textbf{Features}\\
        \hline
      XGB-ProWSyn  &     0.93 &               8.11 &    0.94 &              6.92 &        30 \\\hline
LDA-ProWSyn  &     0.93 &               0.36 &    0.93 &              3.98 &        30 \\\hline
MLP-CCR      &     0.93 &               8.21 &    0.93 &              8.18 &        30 \\\hline
XGB-CCR      &     0.92 &               8.11 &    0.92 &              8.17 &        30 \\\hline
SVC-ClusterCentroids  &     0.91 &               8.18 &    0.91 &              8.26 &        30 \\\hline
LDA-CCR      &     0.93 &               0.54 &    0.93 &              4.06 &        29 \\\hline
SVC-CCR      &     0.90 &              12.93 &    0.90 &             12.89 &        30 \\\hline
MLP-ProWSyn  &     0.87 &              11.45 &    0.90 &              9.38 &        25 \\\hline
SVC-ProWSyn  &     0.90 &               7.13 &    0.90 &              7.10 &        30 \\\hline
KNN-ProWSyn  &     0.90 &              12.75 &    0.90 &             12.90 &        30 \\\hline
LDA-ClusterCentroids  &     0.89 &               8.23 &    0.89 &              8.25 &        28 \\\hline
ET-ClusterCentroids   &     0.89 &               4.27 &    0.89 &              5.67 &        30 \\\hline
MLP-ClusterCentroids  &     0.90 &               8.84 &    0.90 &              9.75 &        29 \\\hline
GNB-CCR      &     0.88 &               5.75 &    0.88 &              5.90 &        30 \\\hline
QDA-ClusterCentroids  &     0.75 &              12.30 &    0.87 &              5.95 &        19 \\\hline
XGB-ClusterCentroids  &     0.90 &               4.37 &    0.90 &              9.66 &        25 \\\hline
KNN-ClusterCentroids  &     0.86 &               9.68 &    0.86 &             10.15 &        26 \\\hline
GNB-ProWSyn  &     0.86 &              11.35 &    0.86 &             11.42 &        28 \\\hline
RF-ClusterCentroids   &     0.86 &               8.41 &    0.86 &              9.71 &        28 \\\hline
QDA-CCR      &     0.72 &               8.66 &    0.83 &              6.30 &        11 \\\hline
QDA-ProWSyn  &     0.51 &               7.35 &    0.82 &             13.70 &         8 \\\hline
GNB-ClusterCentroids  &     0.75 &               8.20 &    0.79 &              6.36 &        15 \\\hline
RF-ProWSyn   &     0.66 &              16.78 &    0.75 &             16.99 &         8 \\\hline
DT-CCR       &     0.64 &              16.36 &    0.74 &             15.81 &        23 \\\hline
ET-ProWSyn   &     0.68 &              11.95 &    0.73 &             14.61 &         6 \\\hline
DT-ClusterCentroids   &     0.72 &              11.02 &    0.72 &             16.14 &        22 \\\hline
AdaB-ClusterCentroids &     0.71 &              16.78 &    0.71 &             14.14 &        21 \\\hline
DT-ProWSyn   &     0.61 &              18.15 &    0.71 &             21.42 &         7 \\\hline
AdaB-ProWSyn &     0.68 &              17.96 &    0.70 &             17.87 &        29 \\\hline
ET-CCR       &     0.68 &              16.04 &    0.68 &             16.19 &        28 \\\hline
KNN-CCR      &     0.63 &              16.35 &    0.67 &              8.45 &        28 \\\hline
AdaB-CCR     &     0.62 &              16.74 &    0.64 &             15.62 &        27 \\\hline
RF-CCR       &     0.57 &              14.47 &    0.59 &             16.67 &        30 \\\hline
        
    \end{tabular}

\captionof{table}{$Acc_{b}$ metric summary - Castellfollit case study. $\overline{Acc_{b}}$ denotes the average value of $Acc_{b}$ of the 10-fold cross validation. Features column display the number of features utilized by each algorithm to achieve this score.}   \label{tbl:models_perfo_castell_app}
\end{table}

\end{document}